\documentclass{article}

\usepackage[nonatbib, final]{neurips_2022}
\usepackage[utf8]{inputenc} 
\usepackage[T1]{fontenc}    
\usepackage[colorlinks]{hyperref}
\usepackage{url}            
\usepackage{booktabs}       
\usepackage{amsfonts}       
\usepackage{nicefrac}       
\usepackage{microtype}      

\usepackage[numbers]{natbib}
\usepackage{graphicx}
\DeclareGraphicsExtensions{.pdf,.png,.jpg,.eps}
\graphicspath{{img/}}

\usepackage{algorithm}
\usepackage[noend]{algpseudocode}	

\usepackage{times}
\usepackage{mathtools}
\usepackage{multirow}
\usepackage{amsthm}
\usepackage{amssymb}
\usepackage{bbm}
\usepackage{bm}
\usepackage{commath}
\usepackage[inline]{enumitem}
\usepackage{caption}
\usepackage{subcaption}

\usepackage[title,titletoc]{appendix}
\usepackage{color}

\usepackage{tikz}
\usepackage{tkz-graph}
\usetikzlibrary{patterns}

\newcommand{\BigO}[1]{\ensuremath{\mathcal{O}(#1)}}                             
\newcommand{\poly}{\mathrm{poly}}
\newcommand{\Norm}[1]{\ensuremath{\lVert #1 \rVert}}                  
\newcommand{\NormI}[1]{\ensuremath{\lVert #1 \rVert}_1}               

\newcommand{\InNorm}[1]{{\left\vert\kern-0.2ex\left\vert\kern-0.2ex\left\vert #1 
    \right\vert\kern-0.2ex\right\vert\kern-0.2ex\right\vert}}                    

\newcommand{\InNormII}[1]{{\left\vert\kern-0.2ex\left\vert\kern-0.2ex\left\vert #1 
    \right\vert\kern-0.2ex\right\vert\kern-0.2ex\right\vert}_2}                    

\newcommand{\InNormInfty}[1]{{\left\vert\kern-0.2ex\left\vert\kern-0.2ex\left\vert #1 
    \right\vert\kern-0.2ex\right\vert\kern-0.2ex\right\vert}_{\infty}}           

\newcommand{\Abs}[1]{\ensuremath{\lvert #1 \rvert}}                              
\newcommand{\iid}{i.i.d.~}                                                        

\newcommand{\defeq}{\overset{\mathrm{def}}{=}}                                   

\DeclareMathOperator*{\union}{\cup}

\newtheorem{definition}{Definition}
\newtheorem{proposition}{Proposition}

\newtheorem{lemma}{Lemma}

\newtheorem{theorem}{Theorem}
\newtheorem{remark}{Remark}

\newtheorem{corollary}{Corollary}

\newcommand{\subto}{\mathrm{subject\ to}}

\def\1{\bm{1}}

\def\rmX{{\mathbf{X}}}

\def\vzero{{\bm{0}}}

\def\vu{{\bm{u}}}
\def\vv{{\bm{v}}}

\def\vx{{\bm{x}}}

\def\mX{{\bm{X}}}

\DeclareMathAlphabet{\mathsfit}{\encodingdefault}{\sfdefault}{m}{sl}
\SetMathAlphabet{\mathsfit}{bold}{\encodingdefault}{\sfdefault}{bx}{n}

\def\gF{{\mathcal{F}}}

\def\gN{{\mathcal{N}}}

\def\gX{{\mathcal{X}}}

\def\sD{{\mathbb{D}}}

\def\sP{{\mathbb{P}}}

\def\sR{{\mathbb{R}}}

\def\sW{{\mathbb{W}}}

\DeclareMathOperator*{\argmin}{arg\,min}

\DeclareMathOperator{\sign}{sign}
\DeclareMathOperator{\Tr}{Tr}

\newcommand{\apptitle}[1]{
\def\toptitlebar{
	\hrule height4pt
	\vskip .25in}

\def\bottomtitlebar{
	\vskip .25in
	\hrule height1pt
	\vskip .25in}

\thispagestyle{empty}
\hsize\textwidth
\linewidth\hsize \toptitlebar {\centering
{\large\bf SUPPLEMENTARY MATERIAL \\ #1 \par}}
\vspace{-0.1in} \bottomtitlebar
}

\newcommand{\Erdos}{Erdős} 
\newcommand{\Renyi}{Rényi}
\newcommand{\had}{\circ}
\newcommand{\kron}{\otimes}
\newcommand{\vect}[1]{\mathrm{vec}(#1)}
\newcommand{\Diag}[1]{\mathrm{Diag}(#1)}
\newcommand{\subjto}{\mathrm{subject\ to}}

\newcommand{\M}{\textsf{M}}

\newcommand{\expm}{\mathrm{expm}}
\newcommand{\hexpm}{h_{\expm}}
\newcommand{\hpoly}{h_{\poly}}
\newcommand{\htinv}{h_{\mathrm{tinv}}}
\newcommand{\hldet}{h_{\mathrm{ldet}}}
\newcommand{\hldetone}{\hldet^{s=1}}
\newcommand{\loss}{\textsf{loss}}
\renewcommand{\sD}{\textsf{DAGs}}
\newcommand{\ftheta}{f_{\theta}}

\renewcommand{\exp}[1]{e^{#1}}

\newcommand{\grad}{\nabla}
\newcommand{\diff}{\partial}

\newenvironment{brsm}
  {\left[\begin{smallmatrix}}
  {\end{smallmatrix}\right]}

\newcommand{\matExample}[4]{\begin{brsm} #1 & #2 \\ #3 & #4 \end{brsm}}

\allowdisplaybreaks 

\author{
	Kevin Bello$^{\dag\ddag}$ \qquad
	Bryon Aragam$^\dag$\qquad
	Pradeep Ravikumar$^\ddag$\\	
	$^\dag$Booth School of Business, University of Chicago, Chicago, IL 60637\\
	$^{\ddag}$Machine Learning Department, Carnegie Mellon University, Pittsburgh, PA 15213
}

\title{DAGMA: Learning DAGs via \M-matrices and a Log-Determinant Acyclicity Characterization}

\begin{document}
\maketitle

\begin{abstract}
The combinatorial problem of learning directed acyclic graphs (DAGs) from data was recently framed as a purely continuous optimization problem by leveraging a differentiable acyclicity characterization of DAGs based on the trace of a matrix exponential function.
Existing acyclicity characterizations are based on the idea that powers of an adjacency matrix contain information about walks and cycles.
In this work, we propose a new acyclicity characterization based on the log-determinant (log-det) function, which leverages the nilpotency property of DAGs. 
To deal with the inherent asymmetries of a DAG, we relate the domain of our log-det characterization to the set of \emph{\M-matrices}, which is a key difference to the classical log-det function defined over the cone of positive definite matrices.
Similar to acyclicity functions previously proposed, our characterization is also exact and differentiable.
However, when compared to existing characterizations, our log-det function: (1) Is better at detecting large cycles; (2) Has better-behaved gradients; and (3) Its runtime is in practice about an order of magnitude faster.
From the optimization side, we drop the typically used augmented Lagrangian scheme and propose DAGMA (\emph{Directed Acyclic Graphs via \M-matrices for Acyclicity}), a method that resembles the central path for barrier methods.
Each point in the central path of DAGMA is a solution to an unconstrained problem regularized by our log-det function, then we show that at the limit of the central path the solution is guaranteed to be a DAG.
Finally, we provide extensive experiments for \emph{linear} and \emph{nonlinear} SEMs and show that our approach can reach large speed-ups and smaller structural Hamming distances against state-of-the-art methods.
Code implementing the proposed method is open-source and publicly available at  \href{https://github.com/kevinsbello/dagma}{https://github.com/kevinsbello/dagma}.
\end{abstract}

\section{Introduction}
\label{sec:introduction}

	Structural equation models (SEMs) \citep{Pearl.2009} are a standard modeling tool in several fields such as economics, social sciences, genetics, and causal inference, to name a few.
	Under this framework, in its general form, the value of each variable in the model is assigned by a general nonlinear, nonparametric function that takes as input the values of other variables in the model, thus, every SEM can be associated to a graphical model.
	In particular, we will consider graphical models that are directed acyclic graphs (DAGs).
	
	A long-standing and active research area deals with the problem of learning the graphical structure (DAG) given passively observed data (a.k.a. causal discovery).
	Computationally, this problem is well-known to be NP-hard in general \citep{chickering1996learning,chickering2004}, mainly due to the combinatorial nature of the space of DAGs.
	In this work, we will follow a score-based approach, a popular learning framework where the goal is to find a DAG that minimizes a \emph{given} score \citep{heckerman1995learning,chickering2002}.
	Recently, \citet{Zheng.2018jsc} proposed an exact smooth nonconvex characterization of acyclicity which opened the door to solving the originally combinatorial problem via a set of tools that work in the fully continuous regime (e.g., gradient-based methods).
	
	Let $W \in \sR^{d\times d}$ be a weighted adjacency matrix of a graph $G$ of $d$ nodes, and let $W \had W$ denote the Hadamard product.
	The acyclicity function introduced by \citet{Zheng.2018jsc} is defined as $\hexpm(W) = \Tr(e^{W\had W}) - d$, where $\Tr$ denotes the trace of a matrix, and it was shown that $\hexpm(W) = 0 $ if and only if $W$ corresponds to a DAG.
	A follow-up work \citep{Yu.2019} proposed another acyclicity function, which can be computed slightly faster, defined as $\hpoly(W) = \Tr((I + \frac{1}{d} W\had W)^d) - d$, where $I$ is the identity matrix.
	It was similarly shown by \citet{Yu.2019} that $\hpoly(W) = 0$ if and only if $W$ corresponds to a DAG.
	While seemingly different, both $\hexpm$ and $\hpoly$ are functions of the form $\Tr(\sum_{k=0}^d c_p (W\had W)^{k}) - d$ for some $c_k > 0$, as noted by \citet{Wei.2020}.
	To the best of our knowledge, all subsequent work \citep[e.g.,][to name a few]{Zheng.2020,lachapelle2019gradient,zhu2020causal,Ng.2020,Ng.2022,moraffah2020causal,kyono2020castle,pamfil2020dynotears}  that has built upon the idea of using a continuous acyclicity characterization have used either $\hexpm$ or $\hpoly$, or some acyclicity characterization in the form of a trace of a sum of matrix powers.
	The latter should come as no surprise, after all, a nonzero diagonal entry of the matrix power $(W\had W)^k$ reveals the existence of a closed walk of length $k$ in $W$.
	
	\textbf{Contributions.}
	In this work, we propose a new acyclicity function that, as $\hexpm$ and $\hpoly$, is both an exact and a smooth acyclicity characterization but that also possesses several advantages when compared to $\hexpm$ and $\hpoly$.
	Specifically, we make the following set of contributions:
	\begin{enumerate}
		\item We propose a novel acyclicity characterization based on the log-determinant (log-det) function (see Theorem \ref{thm:logdet_characterization} and Section \ref{sec:acyclicity_characterization}). In contrast to the classical log-det function defined over the cone of positive definite matrices, we define the domain of our log-det function to be the set of \M-matrices due to the inherent asymmetries of DAGs. To our knowledge, we are the first to connect the notion of \M-matrices to acyclicity and structure learning for DAGs.
		\item We provide a detailed study of the properties of our log-det characterization in Section \ref{sec:properties}. 
			First, we establish the similarities of our log-det function to other existing functions such as $\hexpm$ and $\hpoly$. 
			Second, we formally argue why these functions can be regarded as acyclicity regularizers, similar in spirit to the classical $\ell_1$ and $\ell_2$ regularizers. 
			Third, we show that our log-det function is an \emph{invex} function, i.e., all stationary points are global minimum, moreover, these stationary points correspond to DAGs.
		\item In Section \ref{sec:logdet_better}, we present three arguments as to why our log-det characterization could be preferred over other existing acyclicity functions. Briefly, our log-det function is better at detecting large cycles, has better behaved gradients, and can be computed in about an order of magnitude faster than $\hexpm$ and $\hpoly$.
		\item Motivated by the properties of our log-det function, in Section \ref{sec:optimization}, we present DAGMA (\emph{Directed Acyclic Graphs via \M-matrices for Acyclicity}), a method that resembles the widely known central path approach for barrier methods \citep{Nocedal.20063uv}. We show that, at the limit of the central path, the solution is guaranteed to be a DAG. 
			In contrast to the commonly adopted augmented Lagrangian scheme (originally proposed in \citep{Zheng.2018jsc}) each point in the central path of DAGMA is a solution to an unconstrained problem regularized by our log-det function.
		\item Finally, in Section \ref{sec:experiments} and Appendix \ref{app:experiments}, we provide extensive experiments for \emph{linear} and \emph{nonlinear} SEMs under different score functions (both least squares and log-likelihood), where we show that DAGMA is capable of obtaining DAGs with \emph{better accuracy}, i.e., lower structural Hamming distance (SHD), in a \emph{much faster} way than the state-of-the-art.
			
	\end{enumerate}

\subsection{Related work}
\label{sec:related_work}
	
	The vast majority of methods for learning DAGs can be categorized into two groups: constraint-based algorithms, which rely on conditional independence tests; and score-based algorithms, which focus on finding a DAG that minimizes a given score/loss function.
	We briefly mention classical constraint-based methods as we follow a score-based approach. 
	\citep{spirtes1991} developed the PC algorithm, a popular general method that learns the Markov equivalence class.
	Other algorithms such as \citep{tsamardinos2003algorithms,margaritis1999bayesian} are based on local Markov boundary search. Finally, hybrid approaches that combine constraint-based learning with score-based learning, such as \citep{tsamardinos2006,gamez2011learning}.
	
	In the line of score-based methods, popular score functions include BDeu \citep{heckerman1995learning}, BIC \citep{maxwell1997efficient}, and MDL \citep{bouckaert1993probabilistic}.
	Works that study linear Gaussian SEMs include \citep{Aragam.2019,Aragam.2015,ghoshal2017learning,ghoshal18,Meinshausen.2006,peters2014identifiability}, and for linear non-Gaussian SEMs \citep{Loh.2014,shimizu2006}.
	For nonlinear SEMs, we note works on additive models \citep{buhlmann2014cam,ernest2016causal,voorman2014graph}, additive noise models \citep{hoyer2008nonlinear,Peters.2014,mooij2016distinguishing}, generalized linear models \citep{park2017learning,park2019high,gu2019penalized}, and general nonlinear SEMs \citep{monti2020causal,goudet2018learning}.
	
	More closely related to our work is the line of work built on the nonconvex continuous framework of \citet{Zheng.2018jsc}, such as, \citep{Zheng.2020,lachapelle2019gradient,zhu2020causal,Ng.2020,moraffah2020causal,kyono2020castle,pamfil2020dynotears}.
	In contrast to our work, all of the aforementioned methods rely on the nonconvex acyclicity functions $\hexpm$ or $\hpoly$, and with the exception of \citep{Ng.2020}, all of these works also use the augmented Lagrangian scheme.
	Finally, the NoCurl  method \citep{Yu.2021} also departs from using $\hexpm$, although no other acyclicity constraint is proposed.
	Two immediate distinctions can be made to our work.
	First, we propose a novel acyclicity function based on the log-det function which we show to be prefereable to $\hexpm$ and $\hpoly$.
	Second, we drop the commonly adopted augmented Lagrangian scheme to solve the constrained problem and instead follow a central path approach to leverage the barrier property of our log-det function.
	
	\begin{remark}
		To avoid confusion, we also note that in the GOLEM method of \citet{Ng.2020} the score includes a log-determinant function of the form $\log|\det(I-W)|$ which stems from the \emph{Gaussian log-likelihood}.
		While this expression is zero if $W$ corresponds to a DAG, it is \textbf{not} an exact acyclicity characterization (i.e. $\log|\det(I-W)|=0$ does not imply $W$ is a DAG).
		By contrast, the use of \M-matrices in our work is crucial to translating the log-det function into a valid acyclicity regularizer.
		Moreover, it is not obvious how to extend GOLEM to arbitrary score functions, as their analysis is specific to the Gaussian likelihood function.
	\end{remark}

\section{Notation and background}
\label{sec:background}
	
	\textbf{Notation.}
	We use $[d]$ to denote the set of integers $\{1\ldots d\}$.
	For a square matrix $A$, we use $\lambda_i(A)$ to denote its $i$-th minimum eigenvalue, and use $\rho(A)$ to denote its spectral radius. 
	Also, we use $\Tr(A)$, and $\det(A)$ to denote the trace and determinant of $A$.
	For matrices $A,B$, we let $A\circ B$ represent the element-wise or Hadamard product, moreover, the expression $A \geq B$ is entrywise, i.e., $A_{i,j} \geq B_{i,j}$.
	Then, we say that a matrix $A$ is nonnegative whenever $A \geq 0$.
	For a complex number $a + b i$, we let $\Re(a+bi) = a$ denote its real part.
	We use $\Norm{\cdot}_{p}$ to denote the vector $\ell_p$-norm, and $\Norm{\cdot}_{L^p}$ is the $L^p$-norm on functions.
	Lastly, $i\to j$ and $i \rightsquigarrow j$ represent an edge from $i$ to $j$ and a directed walk from $i$ to $j$, respectively.
	
	Let $X = (X_1,\ldots,X_d)$ be a $d$-dimensional random vector.
	In its general form, a (nonparametric) structural equation model (SEM) consists of a set of equations of the form:
	\begin{align}
	\label{eq:npsem}
	    X_j = f_j(X,Z_j),\ \forall j \in [d],
	\end{align}
	where each $f_j:\sR^{d+1} \to \sR$ is a nonlinear nonparametric function, and $Z_j$ is an exogenous variable representing errors due to omitted factors.
	We consider the Markovian model, which assumes that each $Z_j$ is an independent random variable.
	Note that each $f_j$ depends only on a subset of $X$ (i.e., the parents of $X_j$) and $Z_j$; nonetheless, to simplify notation we ensure that each $f_j$ is defined on the same space.
	Then $f = (f_1,\ldots, f_d)$ induces a graphical structure, where we focus on directed acyclic graphs.
	
	For any joint distribution over $Z=(Z_1,\ldots,Z_d)$, the functions $f_j$ define a joint distribution $\sP(X)$ over the observed data, and a graph $G(f)$ via the dependencies in each $f_j$.
	Then, our goal is to learn $G(f)$ given $n$ \iid samples from $\sP(X)$.
	In score-based learning, given a data matrix $\mX=[\vx_1,\ldots,\vx_d] \in \sR^{n\times d}$, we define a score function $Q(f; \mX)$ to measure the `quality' of a candidate SEM as follows: $Q(f;\mX) = \sum_{j=1}^d \loss(\vx_j,f_j(\mX))$, where we adopt the convention that $f_j(\mX)\in \sR^{n}$.
	Here $\loss$ can be any loss function such as least squares $\loss(\vu,\vv) = \frac{1}{n} \sum_{i=1}^n (u_i-v_i)^2$ or the log-likelihood function, often augmented with a penalty such as BIC or $\ell_1$.
	Given the score function $Q$ and a family of functions $\gF$, we seek to find the $f\in \gF$ that minimizes the score, i.e.,
	\begin{align}
	\label{eq:general_score_based}
		\min_{f\in \gF} Q(f;\mX) \quad \subjto \quad G(f) \in \sD.
	\end{align}
	Similar to \citep{Zheng.2020}, we consider that each $f_j$ lives in a Sobolev space of square-integrable functions whose derivatives are also square integrable.
	Then, let $\partial_k f_j$ denote the partial derivative of $f_j$ w.r.t. $X_k$, it is easy to see that $f_j$ is independent of $X_k$ if and only if $\Norm{\partial_k f_j}_{L^2} = 0$.
	With this observation, we construct the matrix $W(f) \in \sR^{d\times d}$ with entries $[W(f)]_{i,j} \defeq \Norm{\partial_i f_j}_{L^2}$, which precisely encodes the graphical structure amongst the variables $X_j$.
	That is $G(f) \in \sD \iff W(f) \in \sD$, where $W$ is interpreted as the usual weighted adjacency matrix.
	
	In practice, $f$ is replaced with a flexible family of parametrized functions such as deep neural networks, so that problem \eqref{eq:general_score_based} is finite dimensional.
	Finally, note that model \eqref{eq:npsem} includes several models as special cases, e.g., additive noise models, generalized linear models, additive models, polynomial regression, and index models.
	Previous work has studied the identifiability of several of these models, e.g., \citep{hoyer2008nonlinear,shimizu2006,Peters.2014,park2017learning,park2019high,Loh.2014}.
	In the sequel, we assume that the model is chosen such that the graph $G(f)$ is uniquely defined from \eqref{eq:general_score_based}.

\section{A new characterization of acyclicity via log-determinant and \M-matrices}
\label{sec:acyclicity_characterization}

	In this section, we present our acyclicity characterization and study its properties.
	To declutter notation, in this section we simply write $W$ instead of $W(f)$ to denote the \emph{weighted} adjacency matrix of a graph; however, it should be clear that $W$ depends on functions $f_j$ as explained in the previous section.
	
	We develop our characterization by first noting that for any \emph{nonnegative} weighted adjacency matrix $W$, we have that $W \in \sD$ if and only if $W$ is a nilpotent matrix, or equivalently, all the eigenvalues of $W$ are zero, i.e., $\lambda_i(W) = 0, \forall i \in [d]$.
	Then, for any $W \in \sR^{d\times d}$, we have the following obvious implications:
	\begin{align}
		W \in \sD  {\iff} (W\had W) \in \sD    & {\iff}   s - \lambda_i(W\had W) = s, \forall i \in [d], \forall s \in \sR \label{eq:biconditional}\\
		&\hspace{-0.1in}\implies  \prod_{i=1}^d s - \lambda_i(W\had W) = \det(sI - W\had W) = s^d \label{eq:det_relaxation}.
	\end{align}
	Implication \eqref{eq:det_relaxation} can be thought of as a relaxation of acyclicity in the sense that all DAGs satisfy \eqref{eq:det_relaxation}, but not all $W$ that satisfy \eqref{eq:det_relaxation} are DAGs. 
	For example, let $s = 1$ and $W \had W = \matExample{2}{0}{0}{2}$, then it is clear that $\det(sI- W\had W) = 1$ and, thus, \eqref{eq:det_relaxation} is satisfied; however, clearly $W$ is  not a DAG. 
	
	Thus, two immediate questions arise: (i) \emph{Does there exist a domain for $W$ such that \eqref{eq:det_relaxation}$\implies$\eqref{eq:biconditional}?} (ii) \emph{If so, what is the description of such domain?}
	We answer (i) in the affirmative, and answer (ii) by relating the domain of $W$ to the set of \M-matrices, which is defined below.
	
	\begin{definition}[\M-matrix\footnote{More precisely, we consider the definition of a non-singular \M-matrix, which is sufficient for the purposes of this work.}]
	\label{def:m_matrix}
		An \M-matrix is a matrix $A \in \sR^{d\times d}$ of the form $A = sI - B$, where $B \geq 0$ and $s > \rho(B).$
	\end{definition}
	\M-matrices were introduced by \citet{Ostrowski.1937} and arise in a variety of areas including input-output analysis in economics,  linear complementarity problems  in operations  research, finite  difference  methods  for partial  differential  equations,  and Markov chains in stochastic processes.
	To the best of our knowledge, we are the first to connect the notion of \M-matrices to graphical model structure learning through an acyclicity characterization.
	
	The following proposition is an immediate consequence of Definition~\ref{def:m_matrix}:
	\begin{proposition}[\citet{Berman.1994}]
	\label{prop:m_matrix_properties}
		Let $A \in \sR^{d\times d}$ be an \M-matrix, then:
		\begin{center}
			\begin{enumerate*}[label = (\roman*)]
			\item $\Re(\lambda_i(A)) > 0$, for all $i \in [d]$. \label{mmprop:pos_eigs} \qquad\qquad
			\item $A^{-1}$ exists and is nonnegative, i.e., $A^{-1} \geq 0$. \label{mmprop:inv_nonneg}
			\end{enumerate*}
		\end{center}
	\end{proposition}
	
	In the above, item \ref{mmprop:pos_eigs} states that the eigenvalues of an \M-matrix lie in the open right-half plane. Matrices which satisfy the latter property are also known as positive stable matrices. We thus have that \M-matrices are special cases of positive stable matrices.
	It follows that the determinant of any \M-matrix is positive.\footnote{Note that due to asymmetries it is possible for an \M-matrix to have complex eigenvalues. However, since we work with matrices with real entries, the complex eigenvalues come in conjugate pairs.}
	This fact will be used for defining $\hldet^s(W)$, our acyclicity characterization given in Theorem \ref{thm:logdet_characterization}. 
	Finally, the nonnegativity of the inverse from item \ref{mmprop:inv_nonneg} will be used to understand some properties of the gradient of $\hldet^s(W)$.

	We now define the domain over which \eqref{eq:det_relaxation}$\implies$\eqref{eq:biconditional} (see Theorem \ref{thm:logdet_characterization}).
	For any $s>0$, define
    \begin{align}
        \label{eq:defn:Ws}
        \sW^s = \{W\in \sR^{d\times d}\mid s > \rho(W\had W)\},
    \end{align}
    i.e., $\sW^s$ is the set of real matrices whose entry-wise square given by $W\had W$ have spectral radius less than $s$.
	The following lemma lists some relevant properties of $\sW^s$.

	\begin{lemma}
	\label{lemma:properties}
        Let $\sW^s$ defined as in \eqref{eq:defn:Ws}.
		Then, for all $s > 0$:
		\begin{center}
		\begin{enumerate*}[label = (\roman*)]
			\item $\sD \subset \sW^s$. \qquad\qquad
			\item $\sW^s$ is path-connected. \qquad\qquad
			\item $\sW^s \subset \sW^t$ for any $t > s$.
		\end{enumerate*}
		\end{center}
	\end{lemma}
	In Lemma \ref{lemma:properties}, item (i) implies that the $W$ we look for is in the interior of $\sW^s$; item (ii) indicates that we can find a path from any point in $\sW^s$ to any DAG without leaving the set $\sW^s$; item (iii) shows that one can vary $s$ to enlarge or shrink the set $\sW^s$.
	
	Having defined the domain set $\sW^s$, we now define our acyclicity characterization $\hldet^s : \sW^s \to \sR$.
	Recall that item \ref{mmprop:pos_eigs} in Proposition \ref{prop:m_matrix_properties} implies that applying the logarithm function to the determinant of an \M-matrix is always well defined, which motivates our following result.
	\begin{theorem}[Log-determinant characterization]
	\label{thm:logdet_characterization}
		Let $s>0$ and let $\hldet^s : \sW^s \to \sR$ be defined as $\hldet^s(W) \defeq -\log \det (sI - W\had W) + d \log s$.
		Then, the following holds:
		\begin{enumerate}[label = (\roman*)]
			\item $\hldet^s(W) \geq 0,$ with $\hldet^s(W) = 0$ \emph{if and only if} $W$ is a DAG. \label{thm_logdet:item1}
			\item $\grad \hldet^s(W) = 2 (sI - W\had W)^{-\top} \had W$, with $\grad \hldet^s(W) = 0$ \emph{if and only if} $W$ is a DAG. \label{thm_logdet:item2}
		\end{enumerate}
	\end{theorem}

	\subsection{Properties of $\hldet^s(W)$}
	\label{sec:properties}
	
		In this section, we list several properties of our acyclicity characterization.
		The first property we discuss is related to the entries of $\grad \hldet^s(W)$.
		\begin{lemma}
		\label{lemma:entries_Gh}
			For all $i,j \in [d]$, $[\grad \hldet^s(W)]_{i,j} = 0$ if and only if $W_{i,j} = 0$ or there is no directed walk from $j$ to $i$.
			Equivalently, $[\grad \hldet^s(W)]_{i,j} \neq 0$ if and only if the edge $i \to j$ is part of some cycle in $W$.
			Finally, whenever $[\grad \hldet^s(W)]_{i,j} \neq 0$, we have that $\sign([\grad \hldet^s(W)]_{i,j}) = \sign(W_{i,j})$.
		\end{lemma}
		The lemma above characterizes the nonzero entries and their signs of the gradient of $\hldet^s$.
		This property formally offers a \emph{regularizer} perspective for $\hldet^s$, which we highlight next.
		\begin{remark}[A regularizer viewpoint]
			The function $\hldet^s$ promotes small parameters values in much the same way that the classical $\ell_1$ and $\ell_2$ regularizers do.
			In contrast to the latter regularizers, $\hldet^s$ will only shrink the value of a parameter $W_{i,j}$ if and only if the edge $(i,j)$ is part of some cycle in $W$, as prescribed by Lemma \ref{lemma:entries_Gh}.
		\end{remark}
		
		Recall that $\hexpm(W) = \Tr(\exp{W\had W}) - d$ and $\hpoly(W) = \Tr((I + \frac{1}{d} W\had W)^d) - d$.
		It was noted by \cite{Wei.2020} that acyclicity characterizations of the form $\Tr(\sum_{p=1}^d c_p (W\had W)^d)$ for $c_p > 0$ also have the property in Lemma \ref{lemma:entries_Gh}.
		This implies that $\hexpm$ and $\hpoly$ hold the property above and can also be interpreted as acyclicity regularizers.
		We note that the interesting part here is that $\hldet^s$ holds this property besides being different in nature to $\hexpm$ and $\hpoly$.
		
		Next, we state an important consequence of Lemma \ref{lemma:entries_Gh}, which is related to the direction of $\grad \hldet^s(W)$.
		\begin{corollary}
		\label{cor:grad_points_interior}
			At any $W \in \sW^s$, the negative gradient $\grad \hldet^s(W)$ points towards the interior of $\sW^s$.
		\end{corollary}
		
		In optimization, the Hessian matrix plays an important role as it contains relevant information about saddle points and local extrema of a function, and is key to Newton-type methods.	
		Another appealing property of $\hldet^s(W)$ is that it has a Hessian described by a simple closed-form expression.
		\begin{lemma}
		\label{lemma:hess}
			The Hessian of $\hldet^s(W)$, which resides in $\sR^{d^2\times d^2}$, is given by:
			\begin{align*}
				\grad^2 \hldet^s(W) &=  4\ \Diag{\vect{W}} (N \kron N^\top)\ \Diag{\vect{W^\top}} K^{dd}  + 2\ \Diag{\vect{N^\top}},
			\end{align*}
			where $N = (sI - W\had W)^{-1}$, $\kron$ denotes the Kronecker product, and $K^{dd}$ is the $d^2 \times d^2$ commutation matrix such that $K^{dd}\ \vect{A} =\vect{A^\top}$, for any $d\times d$ matrix $A$.
		\end{lemma}
		
		Here we note that among $\hexpm, \hpoly$ and $\hldet^s$, only $\hldet^s$ has a \emph{tractable} expression for the Hessian.
		Furthermore, note that $\grad^2 \hldet^s(W)$ is indexed by vertex pairs so that the entry $[\grad^2 \hldet^s(W)]_{(k,l),(p,q)}$ corresponds to the second partial derivative $\frac{\partial^2 \hldet^s}{\partial W_{k,l} \partial W_{p,q}}$.
		Using Lemma \ref{lemma:hess}, we can characterize the nonzero entries, and their signs, of the Hessian of $\hldet^s$.
		\begin{corollary}
		\label{cor:hess_entries}	
			The entries of the Hessian $\grad^2 \hldet^s(W)$ are described as follows:
			\[
				\left[\grad^2 \hldet^s(W)\right]_{(k,l),(p,q)} = 
					\begin{cases}
						4 W_{l,k} N_{k,q} N_{p,l} W_{q,p}  &\text{if}\ \ (k,l) \neq (p,q),\\
						4 (W_{l,k})^2 (N_{k,l})^2 + 2 N_{k,l}  &\text{if}\ \ (k,l) = (p,q),
					\end{cases}
			\]
			where $N = (sI - W\had W)^{-1}$. 
			Moreover, an off-diagonal entry $[\grad^2 \hldet^s(W)]_{(k,l),(p,q)}$ is nonzero if and only if there exists a cycle in $W$ of the form $q\to p \rightsquigarrow l \to k \rightsquigarrow q$, and has a sign equal to $\sign(W_{l,k} W_{q,p})$.
			Lastly, a diagonal entry $[\grad^2 \hldet^s(W)]_{(k,l),(k,l)}$ is nonzero if and only if there exists a directed walk from $k$ to $l$, and its sign is always positive.
		\end{corollary}
		
		Recall from Theorem \ref{thm:logdet_characterization} that all DAGs attain the minimum value and are critical points of $\hldet^s$; thus, DAGs are local (and global) minimum of $\hldet^s$ and the Hessian matrix evaluated at a DAG must be positive (semi)definite.
		Let us corroborate the latter, when $W$ is a DAG, from Corollary \ref{cor:hess_entries} we have that the off-diagonal elements of $\grad^2 \hldet^s(W)$ are zero, while the diagonal entries are nonnegative.
		That is, the Hessian is positive semidefinite whenever $W$ is a DAG. 
		This implies that all its stationary points are global minima: Such functions are called \emph{invex} \citep{Hanson.1981, Martin.1985}.
		
		\begin{corollary}
		\label{cor:invexity}
			Let $s>0$.
			Then, $\hldet^s(W)$ is an invex function, i.e., all its stationary points are global minima, and these correspond to DAGs.
		\end{corollary}
		
		We note that even though $\hexpm$ and $\hpoly$ are also invex functions, they were not explicitly considered as invex functions before. 
		In fact, it was noted in \citep{Zheng.2018jsc} that DAGs were global minima of $\hexpm$ but no characterization of its stationary points were given.
		\citet{Wei.2020} noted that DAGs were stationary points of $\hexpm$ and $\hpoly$ but the notion of invexity was not explicitly stated.
		
		\begin{remark}[A dynamical system perspective]
		\label{remark:dyn_viewpoint}
			The importance of invexity here is that in the eyes of $\hldet^s$, all DAGs are the same.	
			That is, DAGs correspond to the set of attractors in $\hldet^s$, and depending on the initial condition, the system will converge to a different attractor.
			This offers the following viewpoint for the role of the score function $Q(f;\mX)$, namely, ``use the score $Q$ to find a basin of attraction such that the force field of $\hldet^s$ will dictate the trajectory towards a DAG that is equal or close to the ground-truth''.
		\end{remark}
		
\begin{figure}[!ht]
			\centering
			\begin{subfigure}[b]{.28\textwidth}
				\centering
				\includegraphics[width=\textwidth]{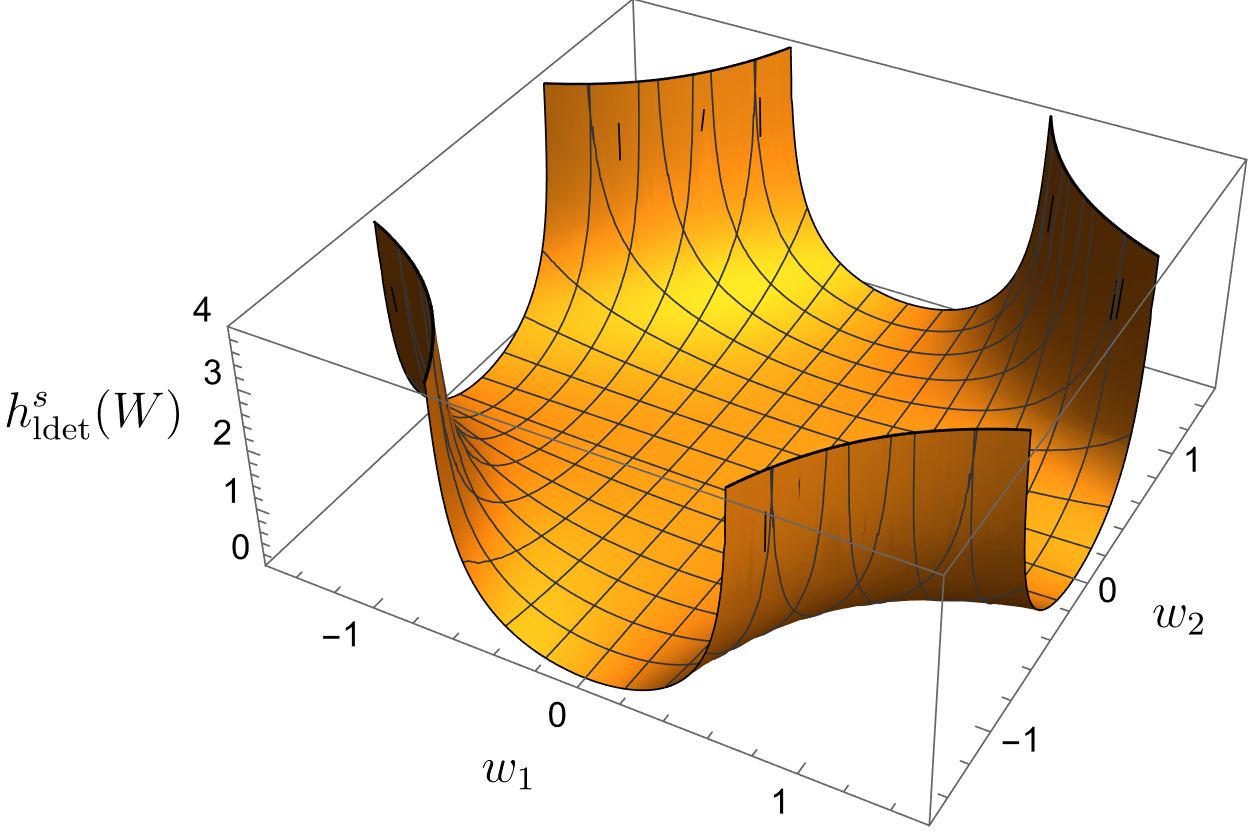}
				\caption{$\hldet^{s=1}(W)$}
			\end{subfigure}
			\qquad
			\begin{subfigure}[b]{.29\textwidth}
				\centering
				\includegraphics[width=\textwidth]{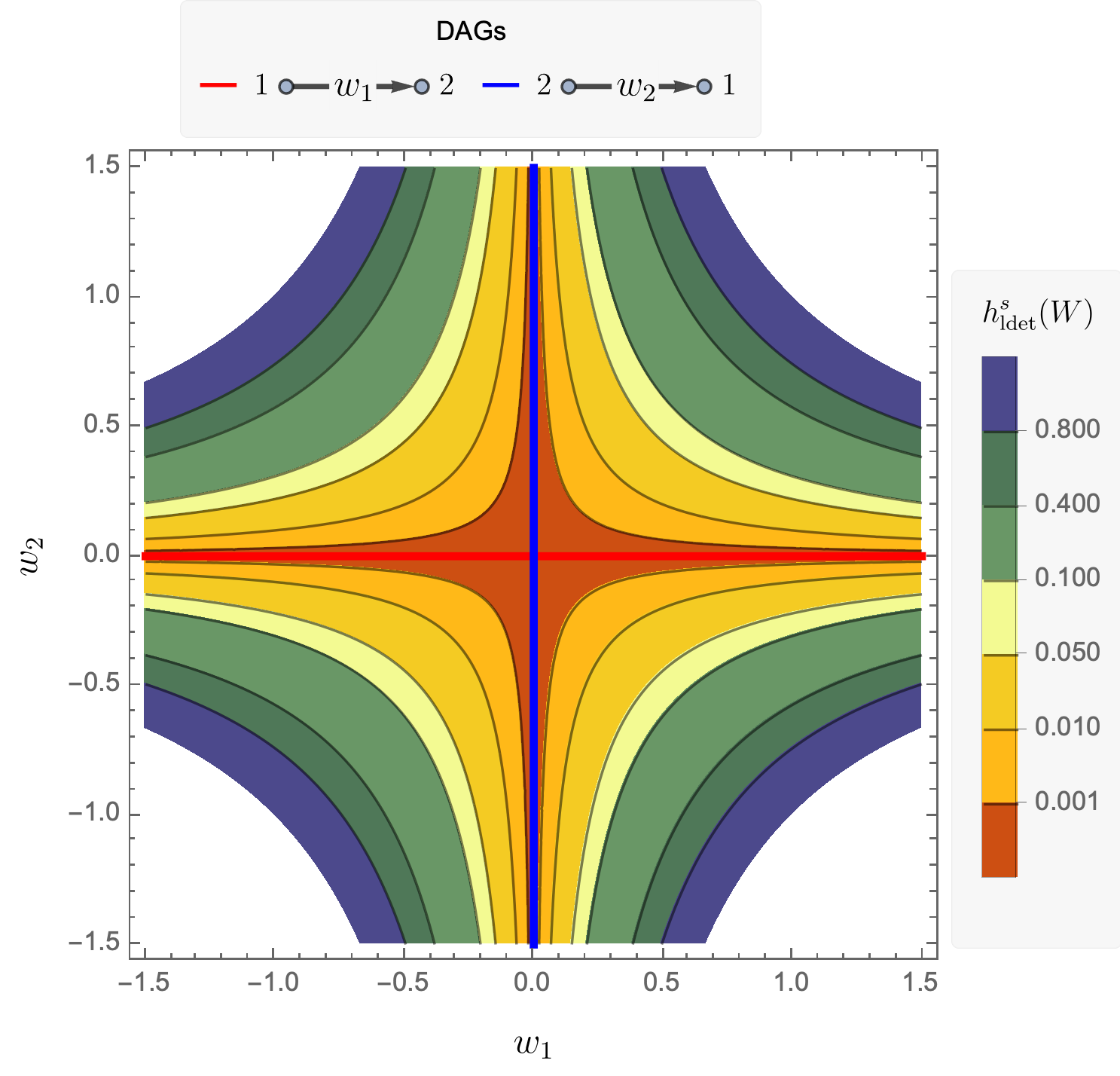}
				\caption{Contours of $\hldet^{s=1}(W)$}
			\end{subfigure}
			\qquad
			\begin{subfigure}[b]{.3\textwidth}
				\centering
				\includegraphics[width=\textwidth]{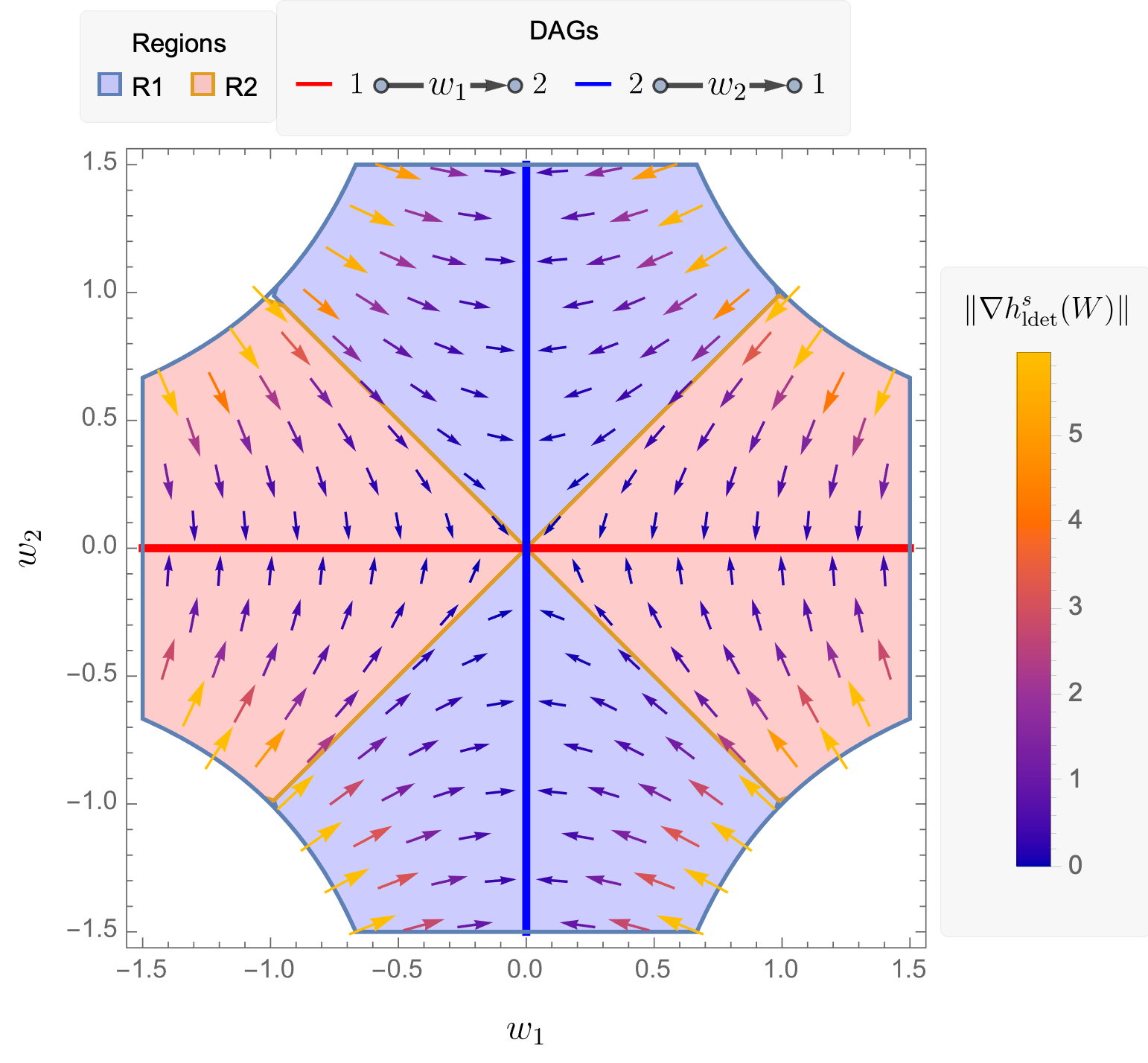}
				\caption{Vector field of $\grad \hldet^{s=1}(W)$}
			\end{subfigure}
			\caption{Behavior of $\hldet^s$ for $W = \matExample{0}{w_1}{w_2}{0}$. Here clearly $W$ is a DAG whenever one of $w_1$ or $w_2$ (or both) are zero.
			In particular, for (c) we note the perspective given in Remark \ref{remark:dyn_viewpoint}, i.e., starting at any point in region R2 will converge to attractors (DAGs) of the form $X_1 \to X_2$ (red line); while starting at any point in region R1 will converge to attractors (DAGs) of the form $X_2 \to X_1$ (blue line).
			}
			\label{fig:h_properties}
		\end{figure}

		In Figure \ref{fig:h_properties}, we illustrate in a toy example the properties discussed in this subsection.

		\subsection{Why the log-determinant regularizer is preferable to existing acyclicity regularizers}
		\label{sec:logdet_better}
			In this section, we present three arguments as to why one should use $\hldet^s$ instead of existing functions such as $\hexpm$ and $\hpoly$.
			We invite the reader to look at Appendix \ref{app:add_discussion} for additional details.
			\begin{enumerate}[wide, label=\bfseries Argument (\roman*).]
			\item \textbf{$\hldet^{s}$ does not diminish cycles of any length.} \label{item:argument_1}
				Let us expand the functions $\hexpm$ and $\hpoly$ in their sum of matrix powers form, that is, $\hexpm(W) = \sum_{k=0}^\infty \nicefrac{1}{k!}\Tr((W\had W)^k) - d$ and $\hpoly(W) = \sum_{k=0}^d  \nicefrac{{d \choose k}}{d^k} \Tr((W\had W)^k) - d$.
				Recall also that the entry $[(W\had W)^k]_{i,i}$ represents the sum of weighted walks from node $i$ to node $i$ of length $k$, where each edge has weight $w^2_{u,v}$.
				Thus, one can notice that if $W$ has cycles of length $k$, their contribution to $\hexpm$ and $\hpoly$ are \emph{diminished} by $\nicefrac{1}{k!}$ and $\nicefrac{{d \choose k}}{d^k}$, respectively.
				Numerically, the latter can be problematic for the following reason: Cycles of length $k$ can go undetected even for small values of $d$ and $k$.
				In practice, a value of $\hexpm, \hpoly \in [10^{-8}, 10^{-5}]$ is typically regarded as zero \citep{Wei.2020,Zheng.2018jsc}.
				Consider a cycle graph of $d$ nodes where each edge weight is $+1$ or $-1$.\footnote{Note that here the sign of an edge is not important since $W\had W$ will have all edge weights equal to $+1$.}
				The plot in Figure \ref{fig:diminishing_cycles} shows how the values of $\hexpm$ and $\hpoly$ decay much faster than that of $\hldet^s$; in fact, at $d = 13$ we already observe $\hexpm(W) \approx 10^{-9}$ and $\hpoly(W) \approx 10^{-14}$, i.e., cycles of length at least $13$ would be numerically undetected by $\hexpm$ and $\hpoly$.
				In contrast, we observe that the value of the log-det function remains bounded away from zero and is able to detect larger cycles.
				\begin{figure}[!ht]
				\begin{center}
					\begin{tikzpicture}[scale=.45,->,>=stealth',auto,node distance=1cm,thick]
					\tikzset{>=latex}
					\tikzstyle{vertex}=[circle, fill=white, draw, inner sep=1pt, minimum size=12pt]
					\node[vertex](1) at (0.5,1) {$1$};
					\node[vertex](2) at (3,1) {$2$};
					\node[] (mid) at (6,1) {$\cdots$};
					\node[vertex](d) at (9,1) {$d$};
					\path[every node/.style={font=\sffamily\small}]
					(1) edge node [right,below] {$1$} (2)
					(2) edge node [right,below] {$-1$} (mid)
					(mid) edge node [right,below] {$1$} (d)
					(d) edge[bend right=30] node [left,above] {$-1$} (1);
					\end{tikzpicture}
				\qquad\qquad
				\includegraphics[width=.4\textwidth]{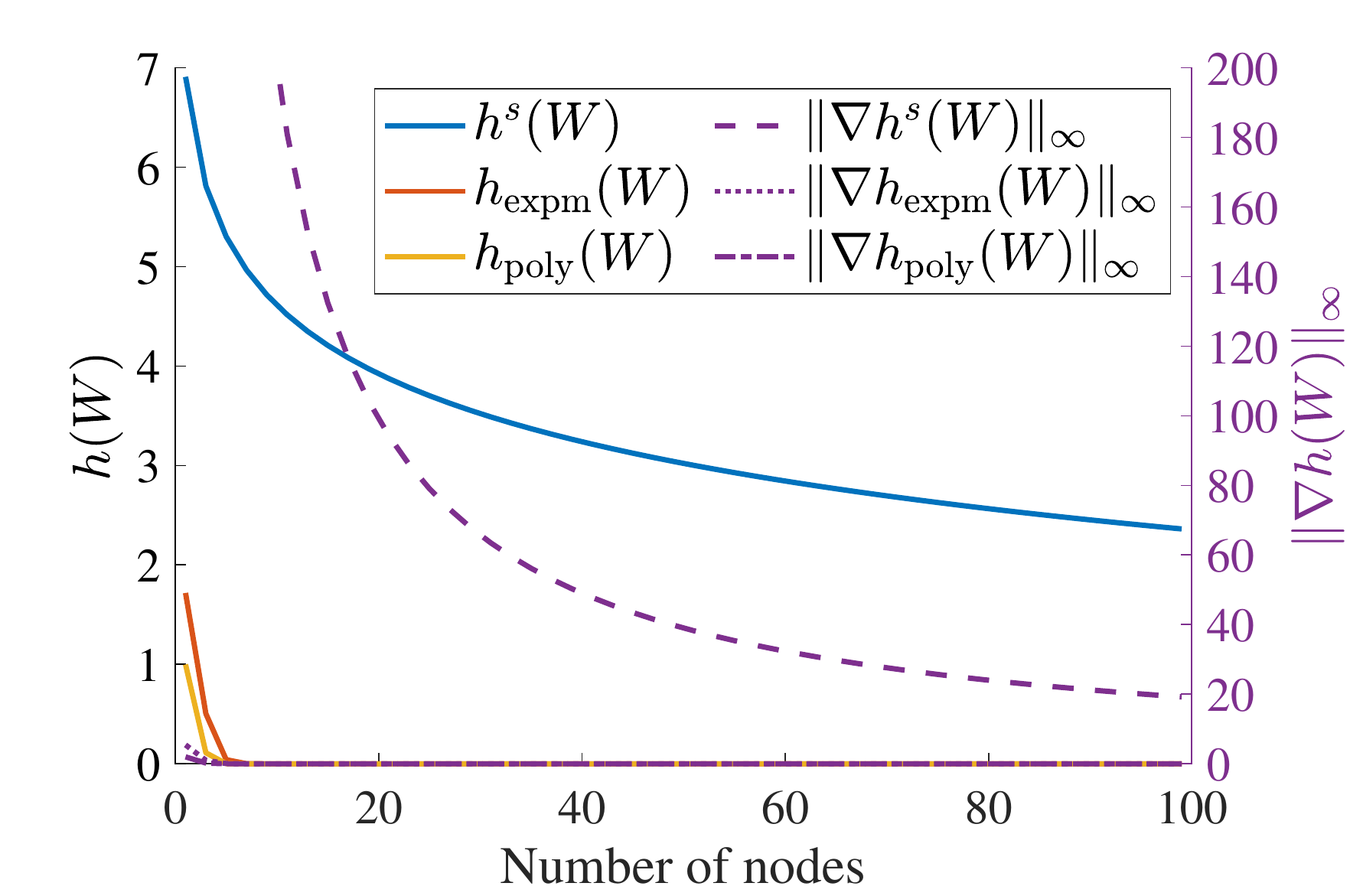}		
				\end{center}
				\caption{
					The values of $\hexpm$ and $\hpoly$ get very close to zero for a number of nodes as small as ten.
					In contrast, letting $s=1.001$, we observe that $\hldet^s$ can stay away from zero even for a cycle graph of 100 nodes.
					Finally, we note a similar pattern for the entrywise $\ell_\infty$ norm of the gradients.}
				\label{fig:diminishing_cycles}
				\end{figure}
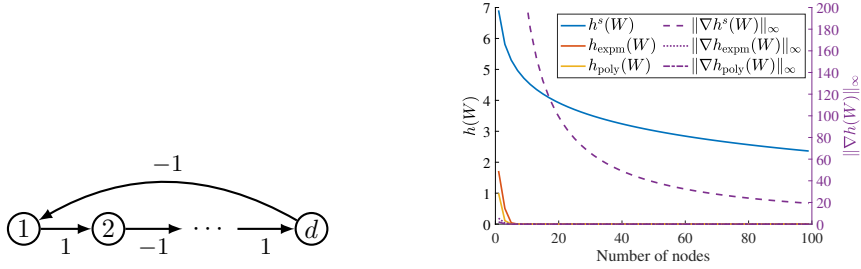
				
				At first glance, it might seem difficult to directly compare the value of $\hldet^s$ to $\hexpm$ and $\hpoly$.
				We next show that when $s=1$, $\hldet^{s=1}$ is an upper bound to $\hexpm$ and $\hpoly$.
				\begin{lemma}
				\label{lemma:upper_bound}
					For all $W \in \sW^{s=1}$, we have $\hpoly(W) \leq \hexpm(W) \leq \hldet^{s=1}(W)$. 
				\end{lemma}
				
				The lemma above shows that in spite of $\hldet^{s=1}$, $\hexpm$, and $\hpoly$ being exact acyclicity characterizations, $\hldet^{s=1}$ will attain the largest value.

			\item \textbf{$\hldet^s$ has better behaved gradients.}
				Similar to argument (i), we show in Appendix \ref{app:add_discussion} that $\hexpm$ and $\hpoly$ are susceptible to vanishing gradients even when the graph contains cycles (see Figure \ref{fig:diminishing_cycles}).
				The following lemma states that the magnitude of each entry of $\grad \hldet^s$ at least as large as the magnitude of the corresponding entry of $\grad \hexpm$ and $\grad \hpoly$, and hence, $\hldet$ has larger gradients to guide optimization.
				\begin{lemma}
				\label{lemma:grad_upper_bound}
					For any walk of length $k$, its contribution to the gradients $\grad \hexpm(W)$ and $\grad \hpoly(W)$ are diminished by $\nicefrac{1}{k!}$ and $\nicefrac{{d-1 \choose k}}{(d-1)^k}$, respectively. 
					In contrast, $\grad\hldet^{s=1}(W)$ does not diminish any walk of any length. 
					This implies that $|\grad\hpoly(W)| \leq |\grad\hexpm(W)| \leq |\grad \hldet^{s=1}(W)|$. 
				\end{lemma}
				
			\item \textbf{Computing $\hldet^s$ and $\grad \hldet^s$ is empirically faster.}
				Even though $\hldet^s$, $\hexpm$, and $\hpoly$ all three share the same computational complexity of $\BigO{d^3}$, in practice $\hldet^s$ can be computed in about an order of magnitude faster than $\hexpm$ and $\hpoly$.
				In Figure \ref{fig:comparison_funcs}, we compare the runtimes of $\hldet^s$, $\hexpm$ and $\hpoly$ for randomly generated matrices, where we observe that computing $\hldet^s$ can be 10x faster than $\hexpm$ and $\hpoly$.
				See Appendix \ref{app:add_discussion} for further details.
				\begin{figure}[!ht]
					\centering
					\includegraphics[width=.5\textwidth]{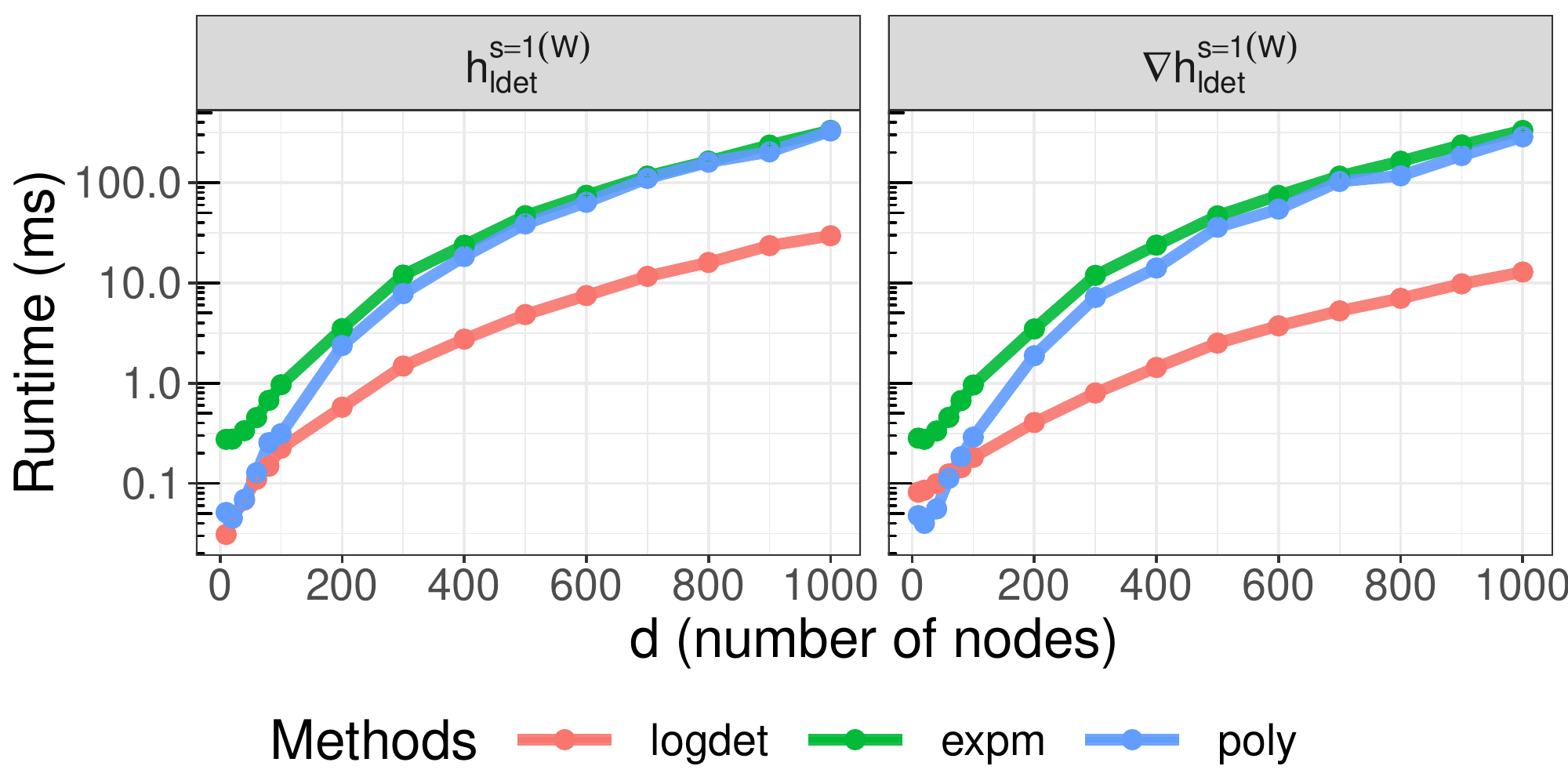}
					\caption{For each $d$, 30 matrices were randomly sampled from a standard Gaussian distribution.}
					\label{fig:comparison_funcs}
				\end{figure}
			\end{enumerate}
		
\vspace{-1em}
\section{Optimization}
\label{sec:optimization}	

	In the previous section we argued why the log-det function should be preferred in practice.
	Let $\ftheta$ denote a  model with parameters $\theta$ for the functions $f_j$ in \eqref{eq:npsem}, e.g., neural networks as in \citep{Zheng.2020}.
	In this section we turn to the problem of minimizing a given score function $Q(\ftheta;\mX)$ constrained to $\hldet^s(W(\theta)) = 0$.
	That is, we aim to solve:
	\begin{align}
	\label{eq:opt_new_acyclicity}
		\min_{\theta} Q(\ftheta;\mX) + \beta_1 \NormI{\theta} \quad \subto\ \ \hldet^s(W(\theta)) = 0,
	\end{align}
	where we include the $\ell_1$ regularizer to promote sparse solutions.
	
	Since the inception of the purely continuous framework for learning DAGs in \citep{Zheng.2018jsc}, almost all follow-up work uses the augmented Lagrangian (AML) scheme to tackle problem \eqref{eq:opt_new_acyclicity}, and L-BFGS-B \citep{Nocedal.20063uv} for solving the sequence of unconstrained problems.
	Motivated by the properties of $\hldet^s$ given in \ref{sec:properties}, we propose a simpler scheme named DAGMA, based on solving a sequence of unconstrained problems in which $\hldet^s$ is simply seen as a regularizer.
	DAGMA resembles the central path approach of barrier methods \citep{Boyd.2004,Nocedal.20063uv}, or the classical path-following approach for solving lasso problems \citep[e.g.][]{friedman2007pathwise}.
	Our method is given in Algorithm \ref{algo:path_following}.
	\begin{algorithm}[!htp]
		\caption{DAGMA}
		\label{algo:path_following}
		\begin{algorithmic}[1]
			\Require{Data matrix $\mX$, initial central path coefficient $\mu^{(0)}$ (e.g., 1), decay factor $\alpha \in (0,1)$ (e.g., $0.1$), $\ell_1$ parameter $\beta_1 > 0$ (e.g., 0.01), log-det parameter $s > 0$ (e.g., 1), number of iterations $T$.}
			\State Initialize $\theta^{(0)}$ so that $W(\theta^{(0)}) \in \sW^s$.
			\For{$t = 0, 1, 2, \ldots T-1$}
				\State Starting at $\theta^{(t)}$, solve $\theta^{(t+1)} = \argmin_{\theta} \mu^{(t)} (Q(\ftheta;\mX) + \beta_1 \NormI{\theta}) + \hldet^s(W(\theta))$
				\State Set $\mu^{(t+1)} = \alpha \mu^{(t)}$
			\EndFor
			\Ensure{$W(\theta^{(T)})$}
		\end{algorithmic}
	\end{algorithm}
	
	The following lemma states that DAGMA will return a DAG at the limit of the central path. This is a critical distinction against existing methods, many of which rely on some type of post-processing (e.g. thresholding) to ensure that the solution is a DAG.
	\begin{lemma}
	\label{lemma:central_path_limit}
		Algorithm \ref{algo:path_following} is guaranteed to return a DAG whenever $\mu^{(t)} \to 0$.
	\end{lemma}

	\subsection{Practical Considerations} 
	\begin{enumerate}
		\item As in barrier methods, where it is required to start at the interior of the feasibility region, in Algorithm \ref{algo:path_following}, we require that the initial point $W(\theta^{(0)})$ be inside $\sW^s$.
		This is very easy to achieve since the zero matrix is in the interior of $\sW^s$ for any $s>0$; therefore, in our experiments we simply set $\theta^{(0)} = 0$.	
		\item Note that in Algorithm \ref{algo:path_following}, we let $\mu^{(t)}$ decrease by a constant factor at each iteration; however, it is possible to specify explicitly the value of each $\mu^{(t)}$, e.g., for $T=4$, we can let $\mu = \{1, 0.1, 0.001, 0\}$.
		\item Regarding the choice of $s$, in principle $s$ could take any value greater than zero since DAGs are inside $\sW^s$ for any $s>0$. 
		Similar to $\mu$, it is also possible to let $s$ vary at each iteration, e.g., for $T=4$, we can set $s = \{1, 0.9, 0.8, 0.8\}$. 
		In practive, we observe that \emph{slightly} decreasing $s$ can help to obtain larger gradients as $W$ gets closer to a DAG.
		Note, however, that letting $s$ be equal or close to 1 is generally easier to optimize than setting $s$ closer to zero, the reason being that for smaller values of $s$ the volume of $\sW^s$ is smaller and will require much smaller learning rates to stay inside $\sW^s$, hence affecting convergence.
		\item Finally, we do not specify how to solve line 3 in Algorithm \ref{algo:path_following}, this is because we leave the door open for different solvers to be used.
		For our experiments in the next section, we solve line 3 by using a first-order method with the ADAM optimizer \citep{kingma2014adam}, which works remarkably well as shown in our experiments. 
		It remains as future work to exploit the Hessian structure of $\hldet$ given in Lemma \ref{lemma:hess} for second-order methods.
	\end{enumerate}

\section{Experiments}
\label{sec:experiments}

	We compare our method against GES~\citep{chickering2002}, PC~\citep{Spirtes.2000}, NOTEARS~\citep{Zheng.2018jsc}, and GOLEM~\citep{Ng.2020} on both \emph{linear} and \emph{nonlinear} SEMs.
	In Appendix \ref{app:experiments}, we specify which existing implementation we used for each of the aforementioned methods.
	Consistent with previous work in this area (e.g., NOTEARS and follow up work), we have not performed any hyperparameter optimization: This is to avoid presenting unintentionally biased results. As a concrete example, for each of the SEM settings, we simply chose a reasonable value for the $\ell_1$ penalty coefficient and used that same value for all graphs across many different numbers of nodes.
	
	Our experimental setting is similar to \citep{Zheng.2018jsc,Zheng.2020}.
	For the main text, we present only a small fraction of all our experiments.
	Moreover, since the accuracy of certain methods were significantly lower than other methods, we report results only against the most competitive ones; full results for all settings and methods can be found in Appendix \ref{app:experiments}. 
	
	\textbf{Linear Models.}
	In Appendix \ref{app:linear_exps}, we report results for linear SEMs with Gaussian, Gumbel, and exponential noises, and use the least squares loss.
	For small to moderate number of nodes, see Appendix \ref{app:linear_small}; for large number of nodes, see Appendix \ref{app:linear_large}; for denser graphs, see Appendix \ref{app:linear_denser}; and for a comparison against GOLEM for sparser graphs, see Appendix \ref{app:dagma_vs_golem}.
	
	\textbf{Nonlinear Models.}
	In Appendix \ref{app:nonlinear_models}, we report results for nonlinear SEMs with binary and continuous data.
	For binary data, we use a logistic model for each structural equation, and use the log-likelihood loss as the score, we report results for small to large number of nodes in Appendix \ref{app:nonlinear_logistic}.
	For continuous data, we consider the continuous additive noise model with Gaussian noise \citep{buhlmann2014cam}, where each nonlinear relationship is modeled by a multilayer perceptron, and use the log-likelihood loss as the score, we report results for small to moderate number of nodes in Appendix \ref{app:nonlinear_mlp}.

\begin{figure}[!ht]
		\centering
		\includegraphics[width=0.8\textwidth]{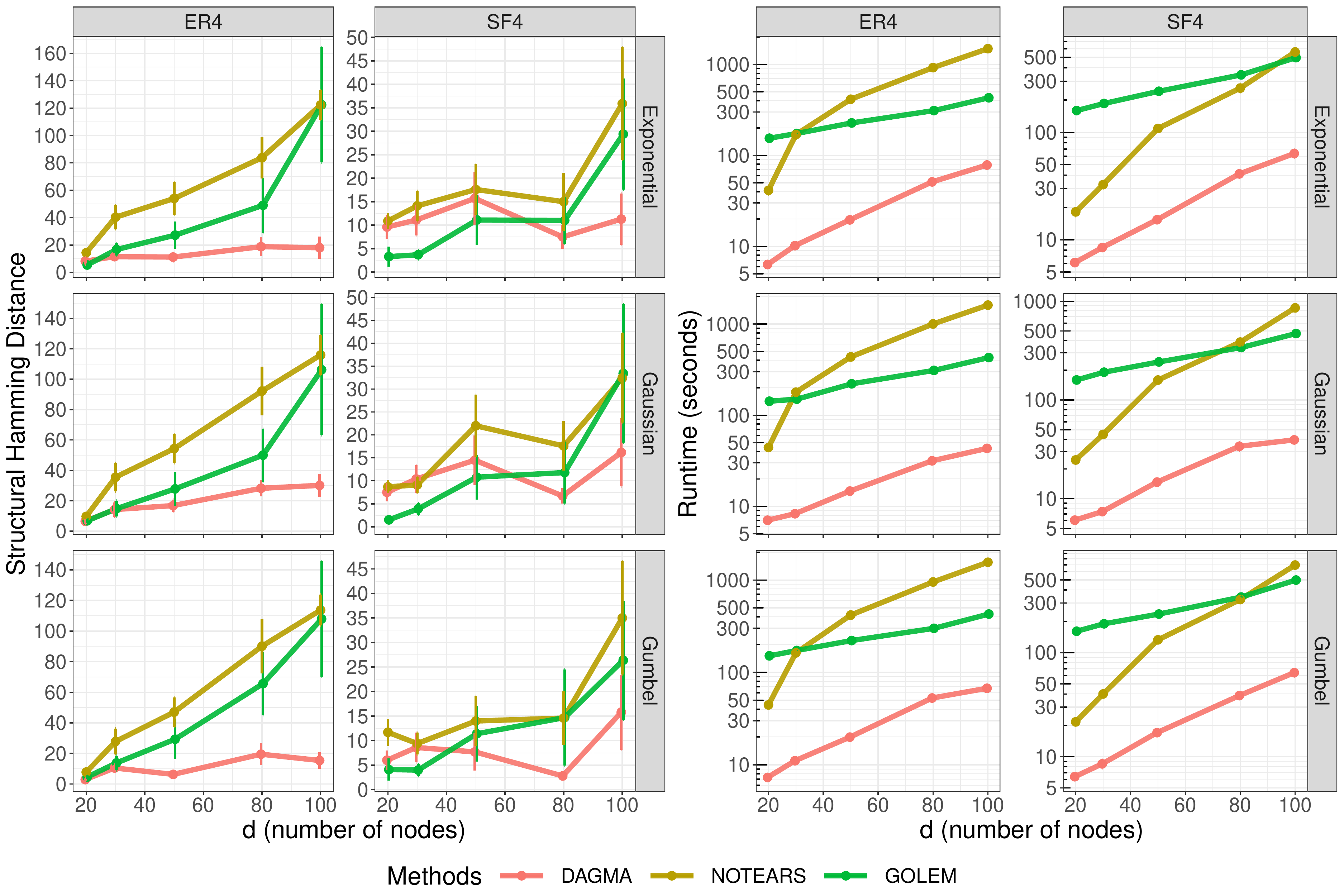}
		\caption{
		Experiments on linear SEMs for $d \in [20,100]$.
		Each point in the plot is estimated over 10 repetitions, where error bars are the standard error. 
		Wall time limit was set to 36 hours.
		}
		\label{fig:main_exps_linear_small}
	\end{figure}

\begin{figure}[!ht]
		\centering
		\includegraphics[width=0.8\textwidth]{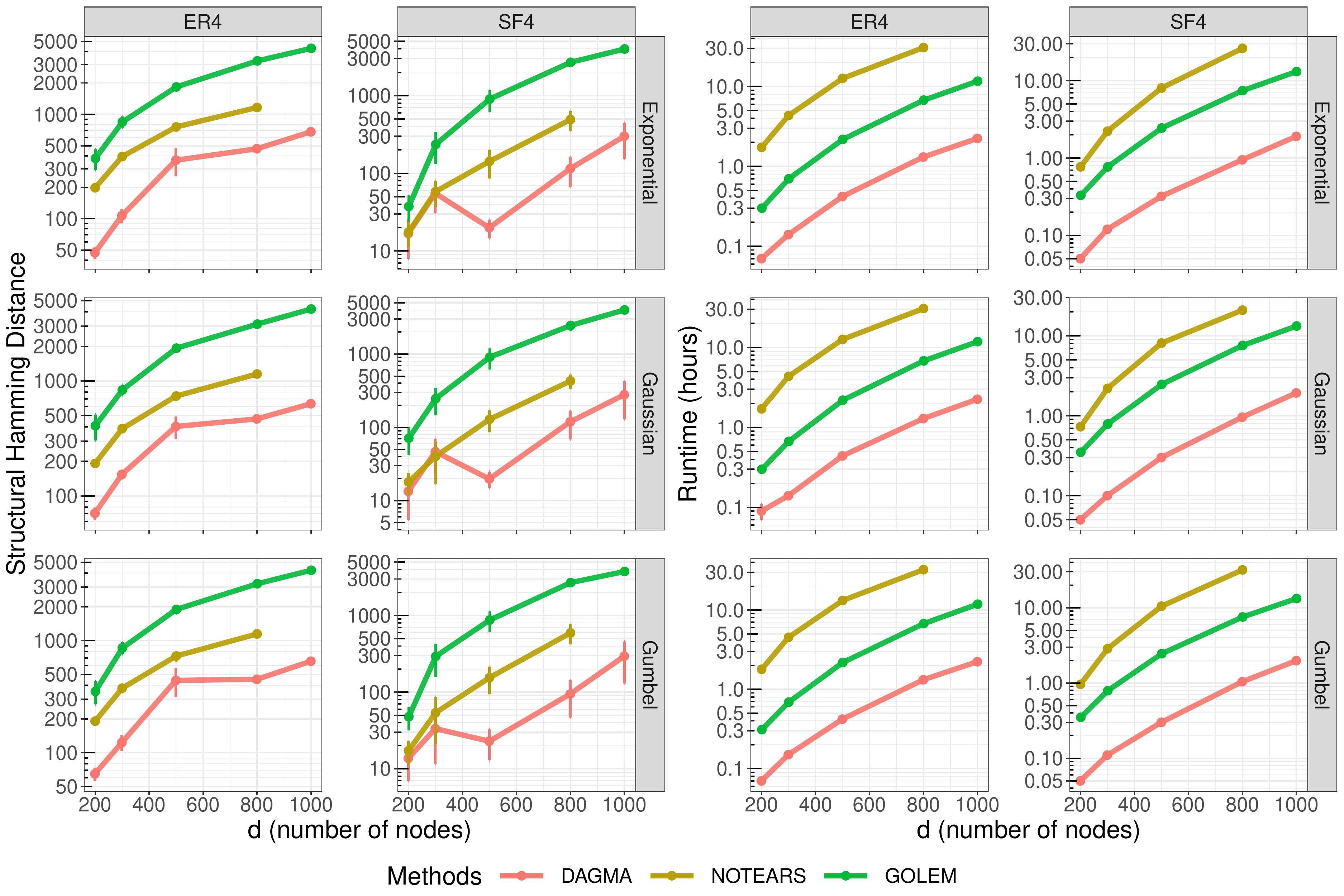}
		\caption{
		Experiments on linear SEMs for $d \in [200,1000]$.
		Each point in the plot is estimated over 10 repetitions, where error bars are the standard error. 
		Wall time limit was set to 36 hours.
		}
		\label{fig:main_exps_linear_large}
	\end{figure}

\begin{figure}[!ht]
		\centering
		\includegraphics[width=0.45\textwidth]{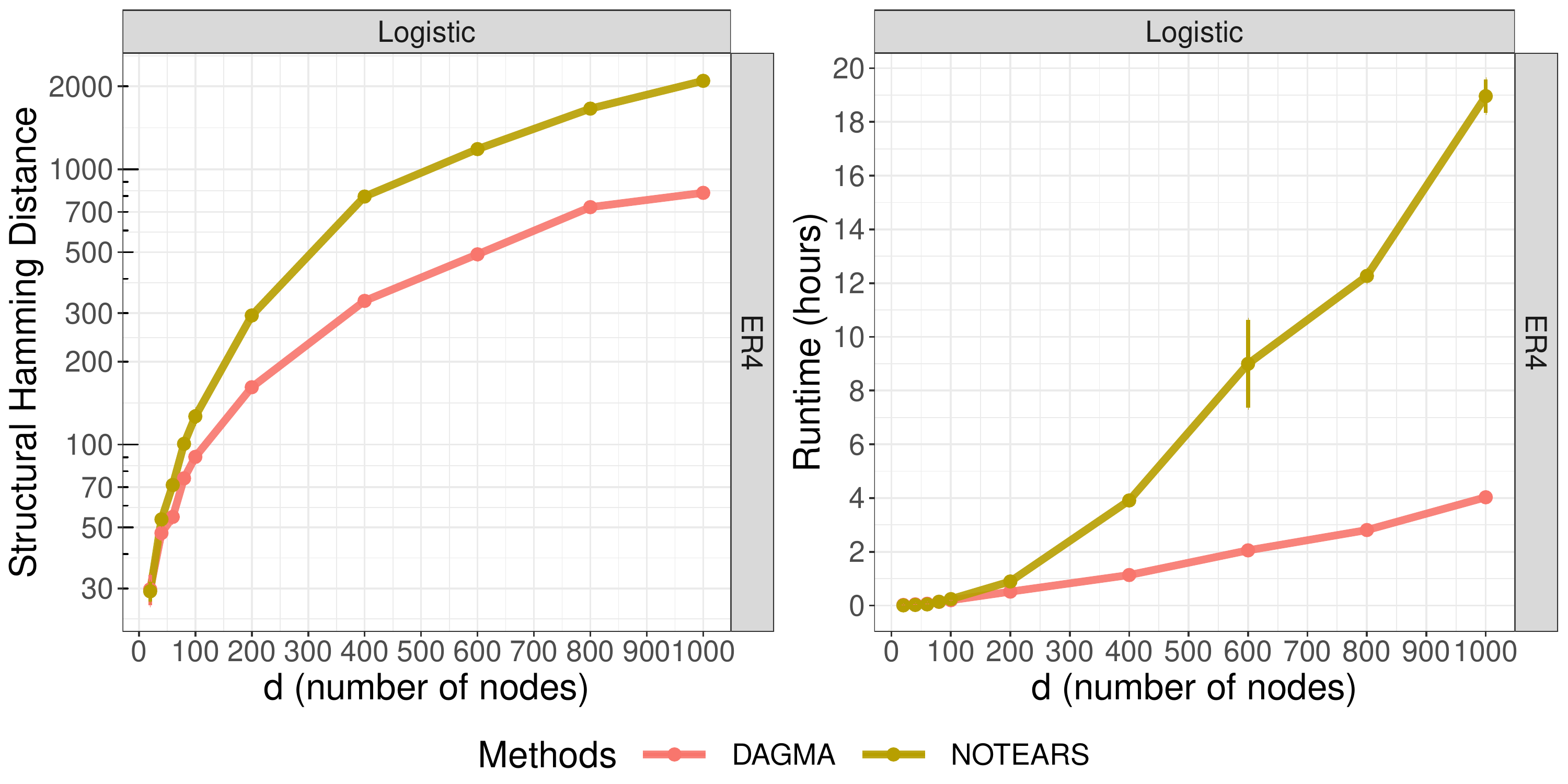}
		\hfill
		\includegraphics[width=0.45\textwidth]{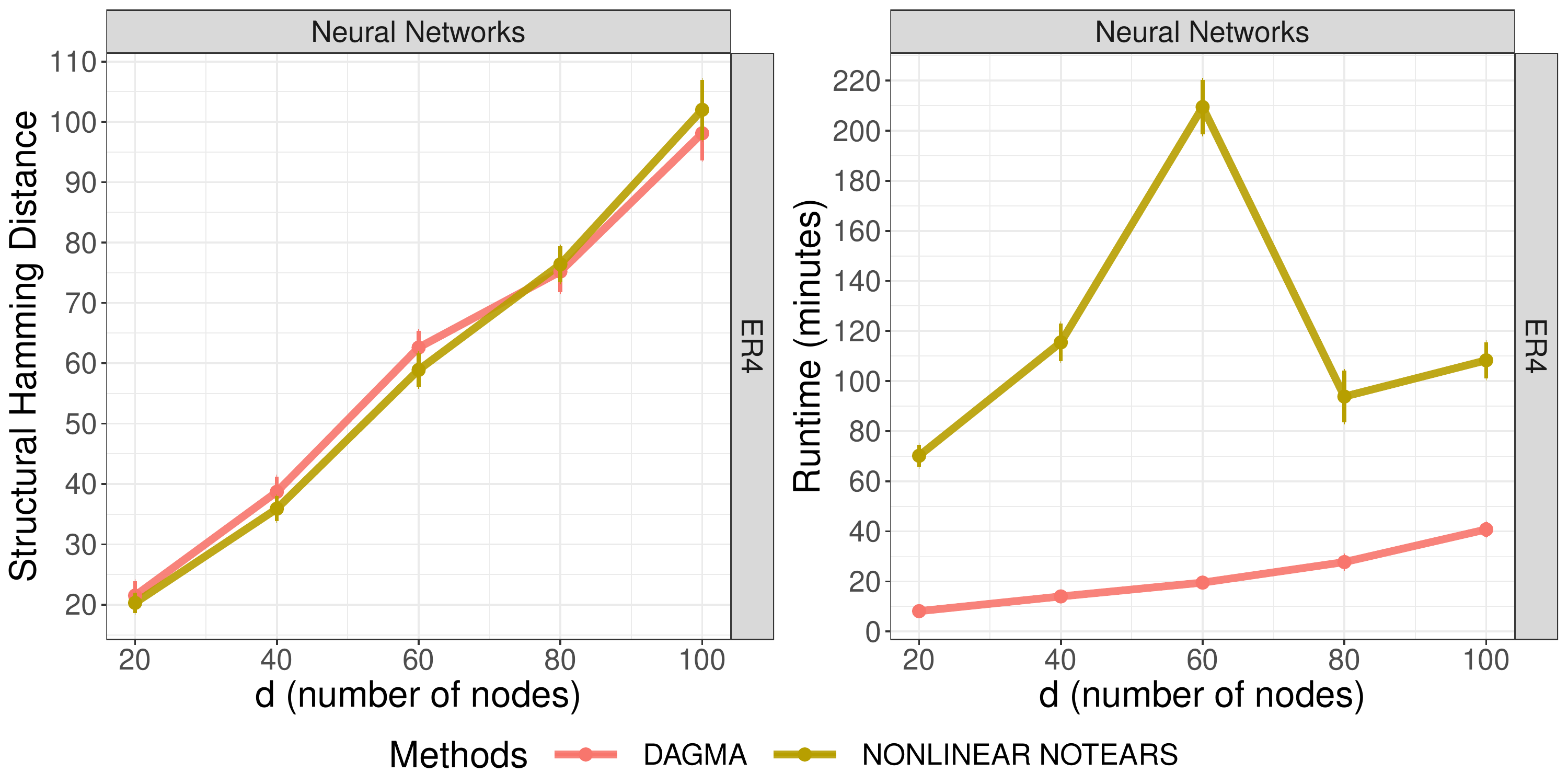}
		\caption{
		Experiments on nonlinear SEMs.
		Each point in the plot is estimated over 10 repetitions, where error bars are the standard error. 
		Wall time limit was set to 36 hours.
		}
		\label{fig:main_exps_nonlinear}
	\end{figure}

    In the following figures, ER4 and SF4 denote \Erdos-\Renyi\ and scale-free graphs, respectively, where for each number of nodes $d$, each graph has in expectation $4d$ edges. 
    It is worth noting that the empirical settings by \citet{Zheng.2018jsc,Zheng.2020} consider graph models such as ER1, ER2, SF1, and SF2.
	Here we focus on the \textbf{hardest} setting, i.e., \textbf{ER4} and \textbf{SF4} graphs.
    For linear SEMs, Figure \ref{fig:main_exps_linear_small} shows results for graphs with $d \in [20,100]$, and  Figure \ref{fig:main_exps_linear_large} shows results for graphs with $d \in [200,1000]$.
    In both regimes, we note that DAGMA obtains significant speedups and improvements in terms of structural accuracy (SHD) against NOTEARS and GOLEM, even though GOLEM is \emph{specific} to and specialized for linear Gaussian SEMs.
    For nonlinear SEMs, Figure \ref{fig:main_exps_nonlinear} shows results for logistic models, we similarly observe that DAGMA attains major speedups and improvements on SHD against NOTEARS \cite{Zheng.2018jsc}.
    Finally, for  nonlinear models using neural networks, we observe that DAGMA is comparable in SHD to the NONLINEAR NOTEARS \citep{Zheng.2020} but obtains significant speedups.
    Again, we invite the reader to look at Appendix \ref{app:experiments} for more details and additional experiments.
    
\vspace{-0.15in}
\section{Final Remarks}
\label{sec:final_remarks}	
	A relevant assumption in this work is that of sufficiency, that is, there are no hidden variables that are a common cause of at least two observed variables.
    While this assumption is widely used for structure learning, we nonetheless highlight that in practice it is very difficult to find scenarios where such assumption holds. As with all work that assumes sufficiency, our work is an important necessary step to understanding settings with hidden variables.
    Finally, we note that the work by \citep{Bhattacharya.2020} proposes a differentiable approach for ADMG for the semi-Markovian case using $\hexpm$.
    It is left for future work to explore the performance of such method using $\hldet$.
    
	Another important limitation of this and previous work on the continuous framework for learning DAGs is that of providing guarantees on the learned structure.
	As in real-life applications one does not have access to the ground-truth DAG, there is much uncertainty as to whether an edge in the predicted DAG actually corresponds to a causal relation.
	Thus, there is still a need for formal guarantees under the continuous framework. 

\vspace{-0.12in}
 \begin{ack}
 \vspace{-0.05in}
 K. B. was supported by NSF under Grant \# 2127309 to the Computing Research Association for the CIFellows 2021 Project.
 B.A. was supported by NSF IIS-1956330, NIH R01GM140467, and the Robert H. Topel Faculty Research Fund at the University of Chicago Booth School of Business.
 P.R. was supported by ONR via N000141812861, and NSF via IIS-1909816, IIS-1955532, IIS-2211907.
 \end{ack}

\bibliographystyle{agsm}
\bibliography{dagma}

\section*{Checklist}

\begin{enumerate}

\item For all authors...
\begin{enumerate}
 \item Do the main claims made in the abstract and introduction accurately reflect the paper's contributions and scope?
    \answerYes{}
 \item Did you describe the limitations of your work?
    \answerYes{(See Section \ref{sec:final_remarks})}
 \item Did you discuss any potential negative societal impacts of your work?
    \answerYes{(See Section \ref{sec:broader_impacts})}
 \item Have you read the ethics review guidelines and ensured that your paper conforms to them?
    \answerYes{}
\end{enumerate}

\item If you are including theoretical results...
\begin{enumerate}
 \item Did you state the full set of assumptions of all theoretical results?
    \answerYes{(See Section \ref{sec:background} and \ref{sec:acyclicity_characterization})}
        \item Did you include complete proofs of all theoretical results?
    \answerYes{(See Section \ref{app:proofs})}
\end{enumerate}

\item If you ran experiments...
\begin{enumerate}
 \item Did you include the code, data, and instructions needed to reproduce the main experimental results (either in the supplemental material or as a URL)?
    \answerYes{(See Section \ref{app:experiments})}
 \item Did you specify all the training details (e.g., data splits, hyperparameters, how they were chosen)?
    \answerYes{(See Section \ref{app:experiments})}
        \item Did you report error bars (e.g., with respect to the random seed after running experiments multiple times)?
    \answerYes{(See Section \ref{app:experiments})}
        \item Did you include the total amount of compute and the type of resources used (e.g., type of GPUs, internal cluster, or cloud provider)?
    \answerYes{(See Section \ref{app:experiments})}
\end{enumerate}

\item If you are using existing assets (e.g., code, data, models) or curating/releasing new assets...
\begin{enumerate}
 \item If your work uses existing assets, did you cite the creators?
    \answerYes{(See Section \ref{app:experiments})}
 \item Did you mention the license of the assets?
    \answerNA{}
 \item Did you include any new assets either in the supplemental material or as a URL?
    \answerYes{}
 \item Did you discuss whether and how consent was obtained from people whose data you're using/curating?
    \answerNA{}
 \item Did you discuss whether the data you are using/curating contains personally identifiable information or offensive content?
    \answerNA{}
\end{enumerate}

\item If you used crowdsourcing or conducted research with human subjects...
\begin{enumerate}
 \item Did you include the full text of instructions given to participants and screenshots, if applicable?
    \answerNA{}
 \item Did you describe any potential participant risks, with links to Institutional Review Board (IRB) approvals, if applicable?
    \answerNA{}
 \item Did you include the estimated hourly wage paid to participants and the total amount spent on participant compensation?
    \answerNA{}
\end{enumerate}

\end{enumerate}

\appendix	
\onecolumn
\apptitle{DAGMA: Learning DAGs via \M-matrices and a Log-Determinant Acyclicity Characterization}


\section{Detailed Proofs}
\label{app:proofs}

\subsection{Proof of Theorem \ref{thm:logdet_characterization}}
\begin{proof}	
	We first note that for any $s>0$ and matrix $A \in \sR^{d\times d}$ we have $\det(sA) = s^d A.$
	Then, $\log \det (sI - W\had W) -d\log s= \log( s^d \det (I - s^{-1} W\had W)) -d\log s = \log \det (I - s^{-1} W\had W)$.
	Moreover, since $W\in \sW^{s}$, we have that $s > \rho(W\had W)$ or equivalently $1>\rho(s^{-1} W\had W)$.
	Thus, in the sequel of the proof we set $s = 1$ w.l.o.g.
	
	\textbf{Item \ref{thm_logdet:item2}.} 
	The gradient expression follows from standard matrix calculus \citep{Petersen.2012}.
	From Lemma \ref{lemma:entries_Gh}, it follows that $W$ is an stationary point of $\hldetone(W)$, i.e., $\grad \hldetone(W)=0$, if and only if $W$ corresponds to a DAG.
	
	\textbf{Item \ref{thm_logdet:item1}.}	
	From the item above, we characterized the stationary points of $\hldetone$. 
	Moreover, from Proposition \ref{prop:m_matrix_properties}, we know that for any $W \in \sW^s$, the gradient $\grad \hldetone$ is well-defined since $I-W\had W$ is an \M-matrix and, thus, its inverse exists.
	Finally, note that at the boundary of $\sW^s$, we have that $\hldetone(W) \to \infty$.
	From these observations, we have that the global minima of $\hldetone$ must be in the interior of $\sW^s$ and will correspond to the set of stationary points.
	Hence, DAGs are local and global minima of $\hldetone$.
	
	We conclude by noting that if $W$ is a DAG then we have $\det (I - W\had W) = 1$ and the equality $\hldetone(W) = 0$ holds immediately.
	Since DAGs are global minima, this implies that for all $W \in \sW^s$ we have $\hldetone(W) \geq 0$.
\end{proof}

\subsection{Proof of Lemma \ref{lemma:properties}}
\begin{proof}
	\textbf{Item (i).} The proof follows directly from the fact that the weighted adjacency matrix $W$ of any DAG is a nilpotent matrix. Since $W\had W$ is also a nilpotent matrix, its spectral radius is zero, i.e., $\rho(W\had W) = 0$. Thus, for any $s > 0$, if $W$ is a DAG then $W \in \sW^s$.

	\textbf{Item (ii).} Recall that a space $\gX$ is path-connected if, for any two points $x,y \in \gX$, there exists a continuous function (path) $\phi : [0, 1] \to \gX$ such that $\phi(0) = x$ and $\phi(1) = y$. Note that since the $d\times d$ zero matrix, $\vzero$, has spectral radius zero, it clearly follows that $\vzero \in \sW^s$ for any $s>0$. We prove path-connectedness of $\sW^s$ by showing that for any $W \in \sW^s$ there exist a path $\phi$ to $\vzero$. Then, for any $W \in \sW^s$, define $\phi(t) = (1-t)W$. It is clear that $\phi$ is a continuous function on $t$, where $\phi(0) = W \in \sW^s$ and $\phi(1) = 0 \in \sW^s$. Now we need to show that $\phi(t) \in \sW^s$ for all $t \in (0,1)$. 
	  Let $t_1$  be an arbitrary number in $(0,1)$, by the nonnegativity of $W\had W$ and by $(1-t_1)^2 < 1$, we have that $(1-t_1)^2 W\had W < W\had W$.
	  Finally, by Perron-Frobenius theory on nonnegative matrices, we have that $\rho\left((1-t_1)^2 W\had W\right) < \rho\left(W\had W\right) < s$, where the last inequality follows by $W \in \sW^s$, which implies that $\phi(t_1) \in \sW^s$. As the choice of $t_1$ was arbitrary, we conclude the proof.
	
	\textbf{Item (iii).} The proof follows immediately by the definition of $\sW^s$.
\end{proof}

\subsection{Proof of Lemma \ref{lemma:entries_Gh}}
\label{proof:entries_grad}
\begin{proof}
	First, recall that $\grad \hldet^s(W) = 2 (sI - W\had W)^{-\top} \had W$.
	Second, since $(sI-W\had W)$ is an \M-matrix, by Proposition \ref{prop:m_matrix_properties}, we have that $(sI - W\had W)^{-\top} \geq 0$.
	By the latter, whenever $[\grad \hldet^s(W)]_{i,j} \neq 0$, we have that $\sign([\grad \hldet^s(W)]_{i,j}) = \sign(W_{i,j})$.
	Finally, from the series expansion of the inverse we have: 
	\[
		(sI - W\had W)^{-1} = \frac{1}{s} I + \frac{1}{s^2} (W\had W) + \frac{1}{s^3} (W\had W)^2 + \cdots,
	\]
	by taking the transpose, that implies that the $i,j$ entry $[(sI - W\had W)^{-\top}]_{i,j}$ is nonzero if and only if there exists a directed walk from $j$ to $i$.
	By taking the Hadamard product, we have that $[(sI - W\had W)^{-\top} \had W]_{i,j}$ is nonzero if and only if $W_{i,j} \neq 0$ and $[(sI - W\had W)^{-\top}]_{i,j} \neq 0$, i.e., there must exist a closed walk of the form $i\to j \rightsquigarrow i$. Which concludes the proof.
\end{proof}

\subsection{Proof of Lemma \ref{lemma:hess}}
\begin{proof}
	We use the Magnus-Neudecker convention \citep{Magnus.1985} for calculating the Hessian.
	Then, by taking differentials and vectorizing, we obtain:
	\begin{align}
		\diff (\grad h(W)) &= 2\ \diff N^\top \had W + 2\ N^\top \had \diff W \notag \\
		\vect{\diff (\grad h(W))} &= 2\ \vect{\diff N^\top \had W} + 2\ \vect{N^\top \had \diff W} \notag \\
		&=2\ \Diag{\vect{W}} \vect{\diff N^\top} + 2\ \Diag{\vect{N^\top}}\vect{\diff W}. \label{eq:hessian_almost_finished}
	\end{align}
	Recall that $N = (sI - W\had W)^{-1}$, we now derive the expression for $\vect{\diff N^\top}$,
	\begin{align*}
		\vect{\diff N^\top} &= - (N \kron N^\top)\ \vect{\diff (sI - W\had W)^\top} \\
		&= 2\ (N \kron N^\top)\ \vect{W^\top \had \diff W^\top} \\
		&= 2\ (N \kron N^\top)\ \Diag{\vect{W^\top}} \vect{\diff W^\top} \\
		&= 2\ (N \kron N^\top)\ \Diag{\vect{W^\top}} K^{dd}\ \vect{\diff W},
	\end{align*}
	plugging in the last equality into eq.\eqref{eq:hessian_almost_finished}, we have
	\begin{align*}
		\grad^2 h(W) & =\frac{\diff\ \vect{\grad h(W)}}{\diff\ \vect{ W}}\\
			&= 4\ \Diag{\vect{W}} (N \kron N^\top)\ \Diag{\vect{W^\top}} K^{dd}  + 2\ \Diag{\vect{N^\top}},
	\end{align*}	
	which concludes the proof.
\end{proof}

\subsection{Proof of Lemma \ref{lemma:upper_bound}}
\begin{proof}
	The comparison between $\hexpm$ and $\hpoly$ is straightforward by looking at the coefficients of their series expansions.
	Recall that, $\hexpm(W) = \sum_{k=0}^\infty \nicefrac{1}{k!}\Tr((W\had W)^k) - d$ and $\hpoly(W) = \sum_{k=0}^d  \nicefrac{{d \choose k}}{d^k} \Tr((W\had W)^k) - d$.
	Since $\nicefrac{1}{k!} \geq \nicefrac{{d \choose k}}{d^k}$, it is clear that $\hexpm(W) \geq \hpoly(W)$.
	To prove that $\hldet^{s=1}(W) \geq \hexpm(W)$, we use the fact that every square matrix has a Jordan canonical form.
	Let $W\had W = Q^{-1} J Q$, where $Q$ is an invertible matrix and $J$ is in Jordan normal form (i.e., a block diagonal matrix with $1$s in the super-diagonal), we have that $\Tr(\exp{W \had W}) = \Tr(\exp{J})$.
	Let $\Lambda(W\had W)$  be the set of distinct generalized eigenvalues of $W\had W$, and $m_{\lambda}$ be the multiplicity corresponding to $\lambda \in \Lambda(W\had W)$.
	Then, we have that 
	\[
		\hexpm(W) = \Tr(e^{W\had W}) - d = \sum_{\lambda \in \Lambda(W\had W)} m_\lambda (e^\lambda - 1).
	\]
	From $\hldet^{s=1}(W)$ we have, 
	\[
		\hldetone(W) = -\log\det (I - W\had W) = \sum_{\lambda \in \Lambda(W\had W)} m_\lambda (-\log(1 - \lambda)),
	\]
	where $\log$ denotes the principal branch of the complex logarithm.
	For any complex $\lambda$, we have the Taylor series:
	\begin{align*}
		e^{\lambda} - 1 = \sum_{k=1}^\infty \frac{\lambda^k}{k!}, \qquad 
		-\log(1 - \lambda) = \sum_{k=1}^\infty \frac{\lambda^k}{k},
	\end{align*}
	where both series converges precisely for all complex numbers ${|\lambda| \leq 1, \lambda \neq 1}$, which is the case as $W \in \sW^{s=1}$.
	From the latter, it is clear to see that $\hexpm(W) \leq \hldetone(W)$, which conludes the proof.
\end{proof}

\subsection{Proof of Lemma \ref{lemma:grad_upper_bound}}
\begin{proof}
	First let us write the gradients for the different acyclicity characterizations.
	Then, we have $\grad \hexpm(W) = 2 (e^{W\had W})^\top \had W$, $\grad \hpoly(W) = 2 ((I + \frac{1}{d} W\had W)^{d-1})^{\top} \had W$,  and $\grad \hldetone(W) = 2 ((I - W\had W)^{-1})^{\top} \had W$.
	When taking  absolute values, it is clear that they differ due to the left-hand side of each Hadamard product.
	Thus, we need to look at the entries of: $\Abs{e^{W\had W}}$, $\Abs{(I + \frac{1}{d} W\had W)^{d-1}}$, and $\Abs{(I - W\had W)^{-1}}$.
	From their series expansions we have:
	\begin{align*}
		e^{W\had W} 					&= \sum_{k=0}^\infty \frac{1}{k!} (W\had W)^k, \\
		(I + \frac{1}{d} W\had W)^{d-1} &= \sum_{k=0}^{d-1} \frac{{d-1 \choose k}}{(d-1)^k} (W\had W)^k,\\
		(I - W\had W)^{-1} 				&= \sum_{k=0}^\infty (W\had W)^k.
	\end{align*}
	Since $W\had W$ is nonnegative, each power $(W\had W)^k$ is also nonnegative.
	Therefore, by noting that $\nicefrac{{d-1 \choose k}}{(d-1)^k} \leq \nicefrac{1}{k!} \leq 1$ for all $k$, the statement follows.
\end{proof}

\subsection{Proof of Lemma \ref{lemma:central_path_limit}}
\begin{proof}
	The proof follows by noting that at the limit of the central path ($\mu^{(t)} \to 0$) we solve the following problem:
	\[
		\widehat{\theta} = \argmin_{\theta} \hldet^s(W(\theta)).
	\]
	Then, by the invexity property of $\hldet^s$ (see Corollary \ref{cor:invexity}), it follows that the solution $W(\widehat{\theta})$ must be a DAG.
\end{proof}

\subsection{Proof of Corollary \ref{cor:grad_points_interior}}
\label{proof:grad_points_interior}
\begin{proof}
	For any $W \in \sW^s$, from Lemma \ref{lemma:entries_Gh}, we know that the nonzeros of $\grad \hldet^s(W)$ have the same sign as the corresponding entries in $W$.
	Then, let $Y = W - a \grad \hldet^s(W)$ for a small value $a$ such that $\Abs{Y} \leq \Abs{W}$.
	It follows that $Y \had Y \leq W\had W$. Since $Y\had Y$ and $W\had W$ are nonnegative matrices, by Perron-Frobenius, we have that $\rho(Y\had Y) \leq \rho(W\had W)$.
	The latter implies that $Y \in \sW^s$, thus, the negative gradient, $-\grad \hldet^s$, must point towards the interior of $\sW^s$.
\end{proof}


\subsection{Proof of Corollary \ref{cor:hess_entries}}
\begin{proof}
	Given the Hessian expression in Lemma \ref{lemma:hess}, the expressions for its entries follow by simple algebraic manipulation.
	From the argument in the proof of Lemma \ref{lemma:entries_Gh} (see Apeendix \ref{proof:entries_grad}), we have that $N_{i,j}>0$ if and only if there exist a directed walk from $i$ to $j$.
	By the latter, the signs and cycle interpretations follow.
\end{proof}

\subsection{Proof of Corollary \ref{cor:invexity}}
\begin{proof}
	Follows from Theorem \ref{thm:logdet_characterization}.
\end{proof}

\section{Additional Discussions}
\label{app:add_discussion}

\subsection{Additional Example for Section \ref{sec:properties}}
\begin{figure}[!ht]
	\includegraphics[width=.9\textwidth]{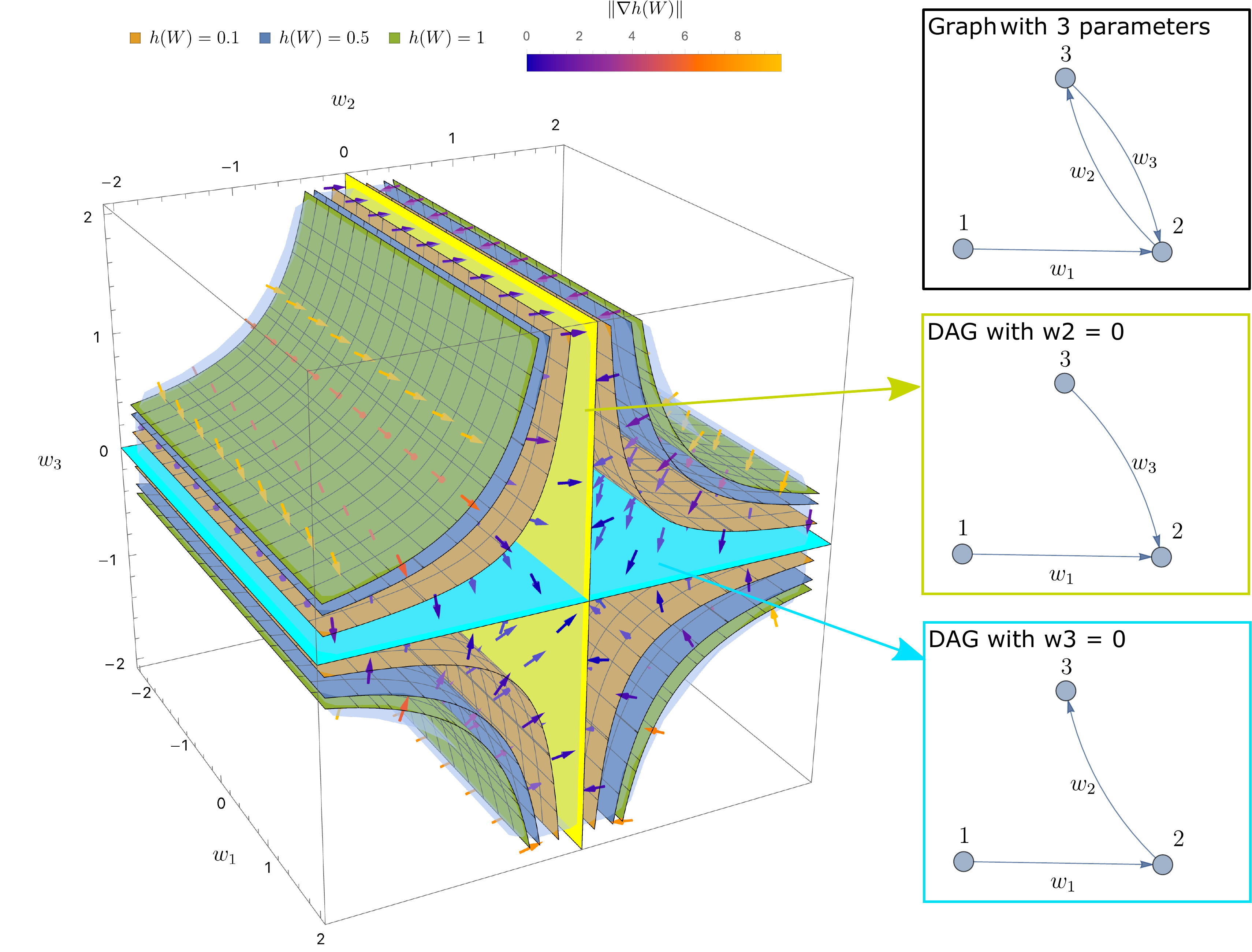}
	\caption{The curved manifolds represent the level sets of $\hldetone$. The arrows represent the vector field $\grad \hldetone$.
	The yellow plane represents DAGs where $w_2 =0$, while the cyan plane represents DAGs where $w_3=0$.
	Similar to Figure \ref{fig:h_properties}, we observe that the negative gradients point towards the interior of $\sW^{s=1}$ and that DAGs are stationary points (attractors) of $\hldetone$.
	}
	\label{fig:supp_3node}
\end{figure}

In Figure \ref{fig:h_properties}, we provided an example of a two-node graph to visualize the properties of $\hldet$.
In Figure \ref{fig:supp_3node}, we present another example for a three-node graph with three edges (parameters).
Specifically, the graph is parameterized by 
$W = \left[\begin{matrix}
	0 & w_1 & 0 \\
	0 & 0 & w_2\\
	0 & w_3 & 0
\end{matrix}\right]$.
Here note that for $W$ to be a DAG at least one of $w_2$ or $w_3$ must be zero.

\subsection{Further Details for Section \ref{sec:logdet_better}}
	In this section, we expand the discussion given in Section \ref{sec:logdet_better} 
	
	\begin{enumerate}[wide, label=\bfseries Argument (\roman*).]
	\item \textbf{$\hldet^{s}$ does not diminish cycles of any length.}
		Let $\htinv(W) = \Tr((I-W\had W)^{-1}) = \sum_{k=0}^\infty \Tr((W\had W)^k) - d$.
		We note that $\htinv$ is another acyclicity characterization, previously considered by \citet{Zheng.2018jsc}.
		From its series expansion, the reader might wonder if simply considering $\htinv(W)$ would be a better alternative than $\hldet(W)$, $\hexpm(W)$, and $\hpoly(W)$.
		The answer is \emph{no}: While this alternative characterization does not suffer from the dimishing cycle problem, it is prone to numerical instability, which was already noted in \citep{Zheng.2018jsc}.
		This numerical instability is mainly due to the exploding gradients of $\Tr((I-W\had W)^{-1})$, which we discuss in our next argument.

	\item \textbf{$\hldet^s$ has better behaved gradients.}
		Let us write the gradients of the different acyclicity characterizations.
		We have:
		\begin{align*}
			&\grad \hexpm(W) = 2 (e^{W\had W})^\top \had W, 
			&\grad \hpoly(W) = 2 ((I + \frac{1}{d} W\had W)^{d-1})^{\top} \had W, \\
			&\grad \htinv(W) = 2 ((I - W\had W)^{-2})^{\top} \had W,
			&\grad \hldet^s(W) = 2 ((sI - W\had W)^{-1})^{\top} \had W.	
		\end{align*}
		From the series expansion for $(I - W\had W)^{-2}$ we obtain:
		\[
			(I - W\had W)^{-2} = \sum_{k=0}^{\infty} (k+1) (W\had W)^{k}.
		\]
		Then, $\grad \htinv$ is numerically unstable whenever $W\had W$ has walks of moderate weight since all walks of length $k$ are now weighted by $k+1$, thus, being prone to exploding gradients.
		Similar to argument (i), we have the following series expansions: 
		\[e^{W\had W} = \sum_{k=0}^{\infty} \frac{1}{k!}(W\had W)^{k},\qquad (I + \frac{1}{d} W\had W)^{d-1} = \sum_{k=0}^{d-1} \frac{{d-1 \choose k}}{(d-1)^k} (W\had W)^{k},\] 
		and
		\[(sI - W\had W)^{-1} = \sum_{k=0}^{\infty} \frac{1}{s^{k+1}}(W\had W)^{k}.\]
		One can observe that for $s=1$, the gradient of the log-determinant weights equally all cycles of any length, whereas $\grad \hexpm$ and $\grad \hpoly$ are again susceptible to the vanishing cycle problem and, thus, they might suffer from vanishing gradients.
		The plot in Figure \ref{fig:diminishing_cycles} shows how the gradients of $\hexpm$ and $\hpoly$ decay at a very fast rate as the cycle graph has more nodes.
		
	\item \textbf{Computing $\hldet^s$ and $\grad \hldet^s$ is empirically faster.}
		In Figure \ref{fig:comparison_funcs}, we compared the runtimes of $\hldet^s$, $\hexpm$ and $\hpoly$ for randomly generated matrices.
		We used the benchmarking library from PyTorch for better runtime estimates over single threads. 
		Experiments were conducted on an Intel Xeon processor E5 v4 with 2.40 GHz frequency.
			
		The reason that computing $\hldet^s$ and $\grad \hldet^s$ is faster is that it involves computing a log-determinant and a matrix inverse respectively, and both of these problems enjoy the large body of work on optimized libraries for matrix factorizations (e.g., LU decomposition) and linear-system solvers.
		In contrast, computing $\hexpm$ relies on a truncated Taylor series of the exponential whose degree is typically estimated using scaling \citep{higham2005scaling,Al-Mohy.2011}, and this requires several matrix-matrix multiplications. The matrix exponential is also a notoriously tricky object to compute \citep{moler2003nineteen}.
		Similar to $\hexpm$, computing $\hpoly$ also requires several matrix-matrix multiplications and, thus, both attain similar performances. 
	\end{enumerate}

\section{Detailed Experiments}
\label{app:experiments}

\paragraph{Computing.}
All experiments were conducted on an 8-core Intel Xeon processor E5-2680v4 with 2.40 GHz frequency, and 32GB of memory.
Each experiment had a wall time of 36 hours.

\paragraph{Graph Models.}
Each simulation in our experiments samples a graph from two well-known random graph models:
\begin{itemize}
    \item \textbf{Erdos-\Renyi} (ER) graphs: These are random graphs whose edges are added independently with equal probability. 
    		We use the notation ER$k$ to indicate that the graph model is an ER graph with $kd$ edges in expectation.
    \item \textbf{Scale-free} (SF) graphs: These are random graphs simulated according to the preferential attachment process \cite{barabasi1999emergence}. 
   			We use the notation SF$k$ to indicate that the graph model is an SF graph with $kd$ edges in expectation and $\beta=1$, where $\beta$ is the exponent used in the preferential attachment process. 
   			It is worth noting that since we consider directed graphs, this particular model corresponds to Price's model, a classical graph model for the growth of citation networks
\end{itemize}

Note that ER graphs are random \emph{undirected} graphs.
To produce a DAG, we draw a random permutation of numbers from $1$ to $d$ and orient the edges respecting this vertex ordering.
For the case of SF graphs, the edges are oriented each time a new node is attached, thus, the sampled graph is a DAG.
Once the ground-truth DAG is generated, we need to simulate the structural equation model, where we provide experiments for linear and nonlinear SEMs.

\begin{remark}
	It is worth noting that the experimental settings by \citet{Zheng.2018jsc,Zheng.2020} consider graph models such as ER1, ER2, ER4, SF1, SF2, and SF4.
	Here we mainly focus in the \textbf{hardest} settings, that is, \textbf{ER4} and \textbf{SF4}.
\end{remark}

\paragraph{Metrics.}
We evaluate the performance of each algorithm with the following four metrics:
\begin{itemize}
    \item \textbf{Structure Hamming distance (SHD)}: A standard measurement for structure learning that counts the total number of edges additions, deletions, and reversals needed to convert the estimated graph into the true graph. 
    \item \textbf{True Positive Rate (TPR)}: Measures the proportion of \emph{correctly} identified edges with respect to the total number of edges in the ground-truth DAG.
    \item \textbf{False Positive Rate (FPR)}: Measures the proportion of \emph{incorrectly} identified edges with respect to the total number of \emph{absent} edges in the ground-truth DAG.
    \item \textbf{Runtime}: Measures how much time the algorithm takes to run, we use it to measure the speed of the algorithms.
\end{itemize}

\begin{remark}
	Consistent with previous work in this area (e.g. NOTEARS and its follow-ups), we have not performed any hyperparameter optimization: This is to avoid presenting unintentionally biased results. 
	As a concrete example, for each of the following SEM settings, we simply chose a reasonable value for the $\ell_1$ penalty coefficient and used that same value for all ER and SF graphs across many different numbers of nodes.
\end{remark}

\subsection{SEM: Linear Setting}
\label{app:linear_exps}
	
	In the linear case, the functions $f_j$ in \eqref{eq:npsem} are directly parameterized by the weighted adjacency matrix $W$. 
	That is, we have the following set of equations:
	\begin{align*}
		X_j = w_j^\top X + Z_j,
	\end{align*}
	where $W = [w_1 \,|\, \cdots \,|\,w_d] \in \sR^{d\times d}$, and $Z_j \in \sR$ represents the noise.
	Here $W$ encodes the graphical structure, i.e., there is an edge $X_k \to X_j$ if and only if $W_{k,j} \neq 0$.

	Then, given the ground-truth DAG $B \in \{0,1\}^{d\times d}$ from one of the two graph models ER or SF, we assigned edge weights independently from $\mathrm{Unif}\left([-2,-0.5]\ \union\ [0.5,2]\right)$ to obtain a weight matrix $W \in \sR^{d\times d}$. 
	Given $W$, we sampled $X = W^\top X + Z \in \sR^{d}$ according to the following three noise models:
	\begin{itemize}
	    \item \textbf{Gaussian noise:} $Z_j \sim \gN(0, 1), \forall j \in [d]$.
	    \item  \textbf{Exponential noise:} $Z_j \sim$ $\mathrm{Exp}(1), \forall j \in [d]$.
	    \item  \textbf{Gumbel noise:} $Z_j \sim$ $\mathrm{Gumbel}(0,1), \forall j \in [d]$.
	\end{itemize}
	Based on these models, we generated random datasets $\mX \in \sR^{n\times d}$ by generating the rows \iid according to one of the models above. 
	For each simulation, we generated $n = 1000$ samples, unless otherwise stated.
	
	To measure the quality of a model, we use the least-square loss 
	\begin{align}
	\label{eq:square_loss}
		Q(W;\rmX)=\frac{1}{2n}\|\rmX-\rmX W\|_F^2,
	\end{align}
	where $\Norm{\cdot}_F$ denotes the Frobenius norm.
	The coefficient $\beta_1$ used for $\ell_1$ regularization, and other parameters required for DAGMA (see Algorithm \ref{algo:path_following}), are later specified for each figure.
	
	The implementation details of the baselines are listed below: 
	\begin{itemize}
	    \item GES (specifically, the FGES algorithm in \citep{ramsey2017}) and PC \citep{Spirtes.2000} are standard baselines for structure learning. Their implementation is based on the \texttt{py-causal} package, available at \url{{https://github.com/bd2kccd/py-causal}}. The exact set of hyperparameters used are:
	    \begin{itemize}
	    	\item For PC: \texttt{testId = `fisher-z-test', depth = 3, fasRule = 2, dataType = `continuous',  conflictRule = 1, 
        		concurrentFAS = True,
 		       useMaxPOrientationHeuristic = True}.
	    	\item For GES: \texttt{scoreId = `cg-bic-score', maxDegree = 5, dataType = `continuous', \\ faithfulnessAssumed = False}.
	    \end{itemize}
	    \item The NOTEARS method in \citet{Zheng.2018jsc} was implemented using the author's Python code available at: \url{https://github.com/xunzheng/notears}. Its score function is also the least square as defined in eq.\eqref{eq:square_loss}. For the $\ell_1$ coefficient, for a fair comparison, we use the same value used for DAGMA. For the rest of hyperparameters, we use their default values.
	    \item The GOLEM method in \citet{Ng.2020} was implemented using the author's Python and Tensorflow code available at: \url{https://github.com/ignavierng/golem}. Here we use their default set of hyperparameters, that is, $\lambda_1=0.02$ and $\lambda_2 = 5$, for other details of their method we refer the reader to Appendix F of \citet{Ng.2020}. 
	\end{itemize}

	\subsubsection{Small to Moderate Number of Nodes}
	\label{app:linear_small}
		Following the aforementioned  process to generate data, in this section, we test the methods for graphs with number of variables $d \in\{ 20, 30, 50, 80, 100\}$. 
		We use the following setting for DAGMA (Algorithm \ref{algo:path_following}): Number of iterations $T=4$, initial central path coefficient $\mu^{(0)} = 1$, decay factor $\alpha = 0.1$, $\ell_1$ coefficient $\beta_1 = 0.05$, log-det parameter $s=\{1, .9, .8, .7\}$.
		For each problem in line 3 of Algorithm \ref{algo:path_following}, we implement an adaptive gradient method using the ADAM optimizer \citep{kingma2014adam}.
		The hyperparameters for ADAM are: Learning rate of $3\times 10^{-4}$, and $(\beta_1,\beta_2) = (0.99,0.999)$.
		For $t = \{0,1,2\}$, we run ADAM for $2\times 10^4$ iterations or until the loss converges, whichever comes first.
		For $t = 3$, we run ADAM for $7\times 10^4$ iterations or until the loss converges, whichever comes first.
		We consider that the loss converges if the relative error between subsequent iterations is less than $10^{-6}.$
		Finally, as in previous work including the baseline methods \citep{Zheng.2018jsc,Zheng.2020,Ng.2020}, a final thresholding step is performed as it was shown to help reduce the number of false discoveries. For all cases, we use a threshold of $0.3$.

\begin{figure}[!tb]
			\centering
			\includegraphics[width=1\textwidth]{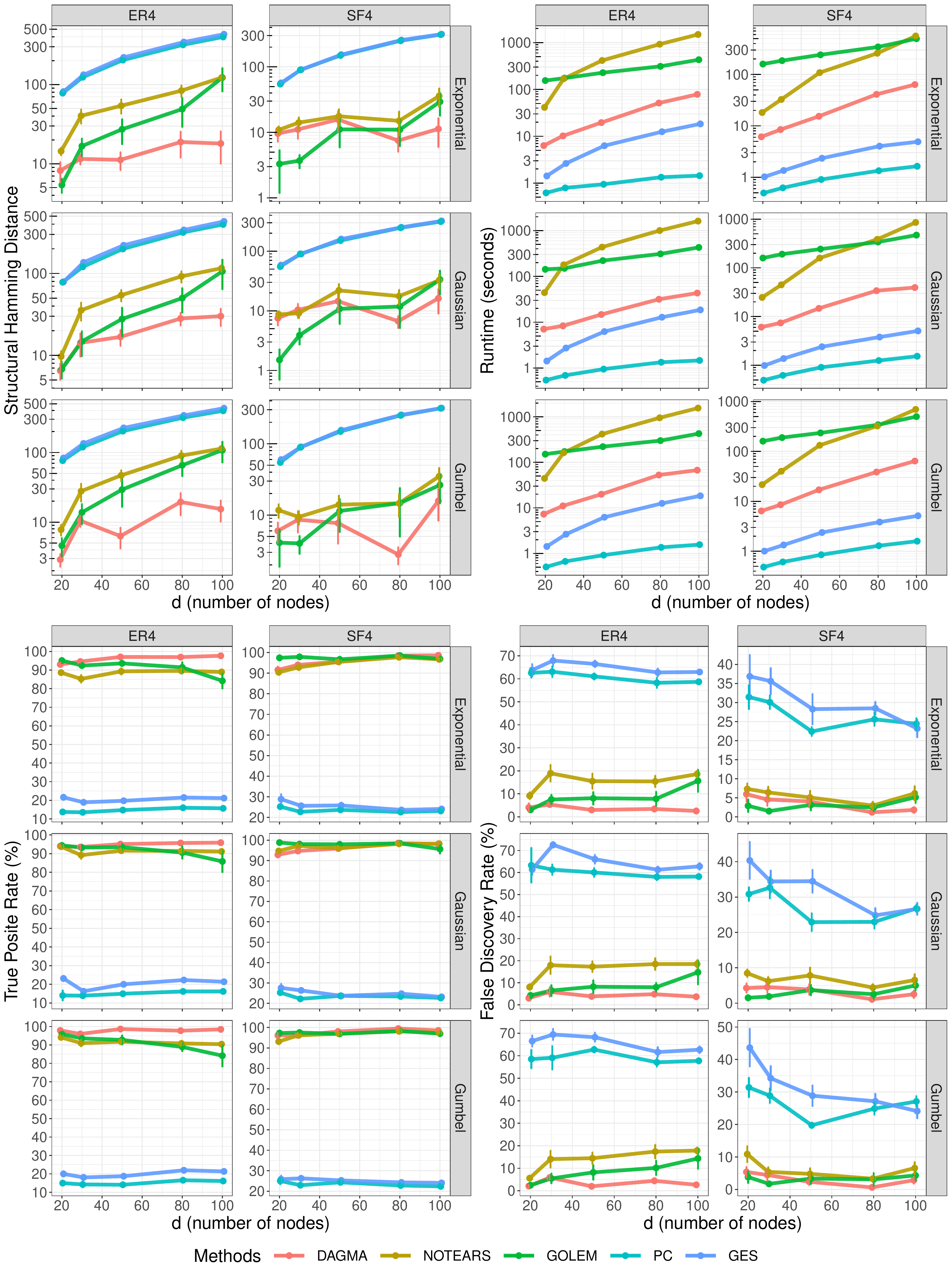}
			\caption{
				SHD, Runtime, TPR and FDR of all methods for different graph types (ER4, SF4) and different noise types (Gaussian, Gumbel, exponential).
				In all cases, lower is better \emph{except} for the TPR.
				Error bars represent standard errors over 10 simulations.
				More details are given in Section \ref{app:linear_small}, and a summary is provided in Table \ref{tab:linear_small}.
			}
			\label{fig:linear_small}
		\end{figure}

		The results for different graph models (ER4, SF4), and different noise distributions (Gaussian, Gumbel, exponential), are shown in Figure \ref{fig:linear_small}.
		In Table \ref{tab:linear_small}, we average the SHDs and runtimes across graph and noise types, for the competitive methods.
		Here we note in particular that for $d=100$, DAGMA obtains an \textbf{improvement} of $74.9\%$ and $76.5\%$ in \textbf{SHD} against GOLEM and NOTEARS, respectively; also, DAGMA runs $7.7$ and $19.1$ times \textbf{faster} than GOLEM and NOTEARS, respectively.
		Finally, we note that DAGMA performs much better than GOLEM besides the latter being \emph{tailored} to linear Gaussian models.	
		
		\begin{table}[!htb]
		\centering
		\caption{
		Summary of performances (SHD and runtime) of the most competitive methods.
		Each metric was averaged across different graph and noise types. 
		Finally, the errors denote $95\%$ confidence intervals on 10 repetitions.
		}
		\label{tab:linear_small}
		\begin{tabular}{@{}cccc@{}}
		\toprule
		\textbf{Method}                   & \textbf{$d$} & \textbf{SHD}    & \textbf{Runtime (seconds)} \\ \midrule
		\multirow{5}{*}{\textbf{DAGMA}}   & 20           & 6.78$\pm$1.64   & 6.54$\pm$ 0.42       \\
		                                  & 30           & 11.05$\pm$2.50  & 8.99$\pm$ 0.62       \\
		                                  & 50           & 12.03$\pm$3.42  & 16.88$\pm$0.98       \\
		                                  & 80           & 13.92$\pm$4.44  & 41.55$\pm$3.30       \\
		                                  & 100          & 17.80$\pm$5.72  & 59.36$\pm$4.80       \\\midrule
		\multirow{5}{*}{\textbf{GOLEM}}   & 20           & 4.28$\pm$1.38   & 154.64$\pm$2.50      \\
		                                  & 30           & 9.48$\pm$3.10   & 177.21$\pm$5.62      \\
		                                  & 50           & 19.60$\pm$7.30  & 231.53$\pm$3.58      \\
		                                  & 80           & 33.68$\pm$12.87 & 324.41$\pm$7.10      \\
		                                  & 100          & 70.95$\pm$26.11 & 458.94$\pm$8.62      \\\midrule
		\multirow{5}{*}{\textbf{NOTEARS}} & 20           & 10.53$\pm$1.58  & 32.41$\pm$3.72       \\
		                                  & 30           & 22.70$\pm$6.04  & 104.76$\pm$20.43     \\
		                                  & 50           & 34.82$\pm$7.96  & 278.13$\pm$40.14     \\
		                                  & 80           & 52.22$\pm$13.49 & 640.95$\pm$89.24     \\
		                                  & 100          & 75.87$\pm$13.97 & 1129.10$\pm$120.74   \\ \bottomrule
		\end{tabular}
		\end{table}
	
	\subsubsection{Large Number of Nodes}
	\label{app:linear_large}
		In this section, we test DAGMA, GOLEM, and NOTEARS for graphs with number of variables $d \in\{ 200, 300, 500, 800, 1000\}$. 
		We do not test PC and GES as they are not competitive in terms of accuracy, as shown in Figure \ref{fig:linear_small}.
		We follow the same setting for DAGMA given in Section \ref{app:linear_small}.
		
\begin{figure}[!tb]
		\centering
		\includegraphics[width=\textwidth]{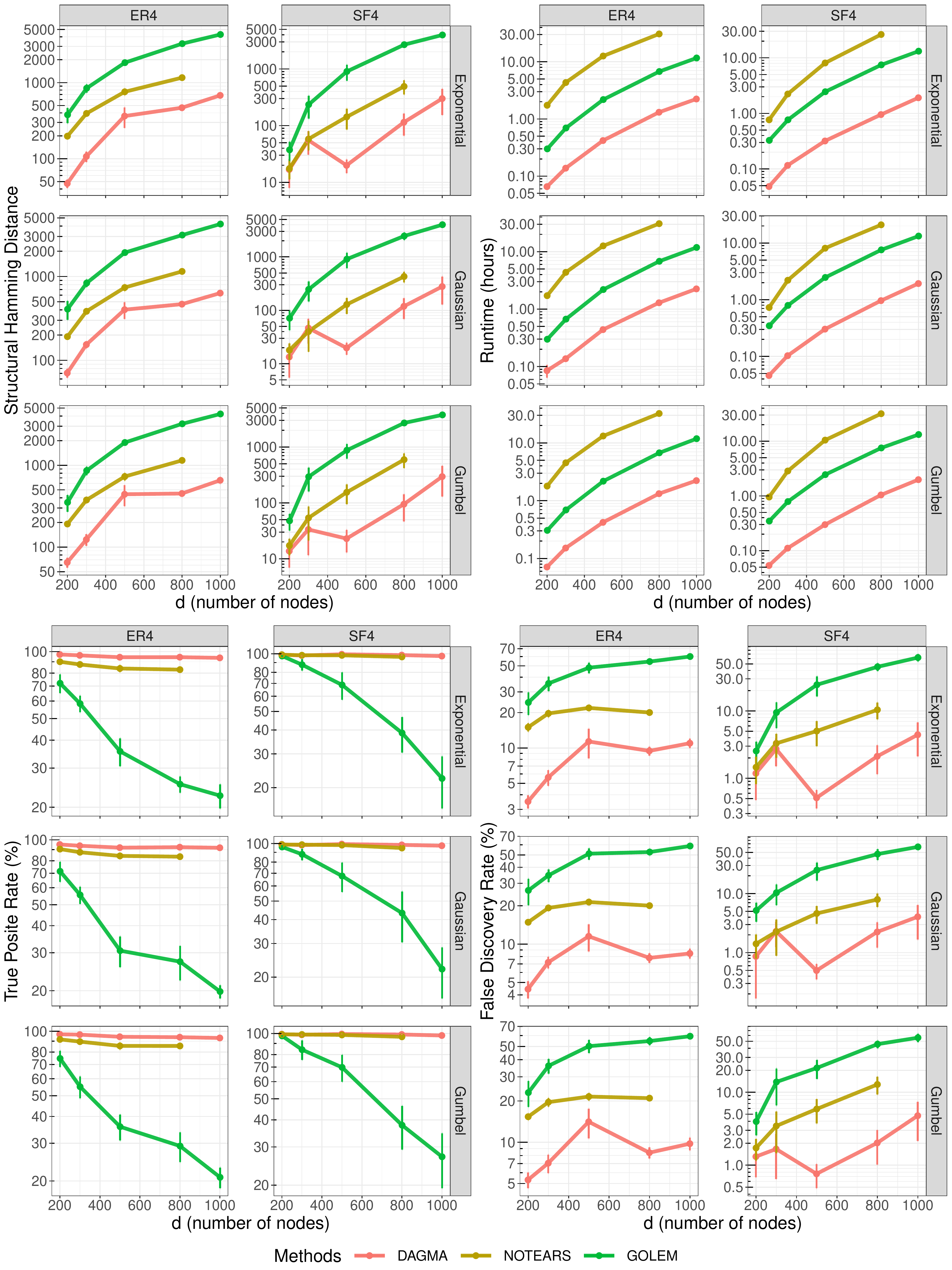}
		\caption{SHD, Runtime, TPR and FDR of competitive methods for different graph types (ER4, SF4) and different noise types (Gaussian, Gumbel, exponential).
				In all cases, lower is better \emph{except} for the TPR.
				Error bars represent standard errors over 10 simulations.
				More details are given in Section \ref{app:linear_large}, and a summary is provided in Table \ref{tab:linear_large}.}
		\label{fig:linear_large}
	\end{figure}

		The results for different graph models (ER4, SF4), and different noise distributions (Gaussian, Gumbel, exponential), are shown in Figure \ref{fig:linear_large}.
		In Table \ref{tab:linear_large}, we average the SHDs and runtimes across graph and noise types.
		Here we note in particular that for $d=800$, DAGMA obtains an \textbf{improvement} of $90.2\%$ and $65.5\%$ in \textbf{SHD} against GOLEM and NOTEARS, respectively; also, DAGMA runs $6.2$ and $25$ times \textbf{faster} than GOLEM and NOTEARS, respectively.
		For $d=1000$, we observe that NOTEARS takes more than 36 hours, which is the reason we could not report its performance.
		Finally, we note that once again DAGMA performs much better than GOLEM besides the latter being \emph{tailored} to linear Gaussian models.
		
		\begin{table}[!ht]
		\centering
		\caption{
		Summary of performances (SHD and runtime) of the most competitive methods.
		Each metric was averaged across different graph and noise types. 
		Finally, the errors denote $95\%$ confidence intervals on 10 repetitions.
		}
		\label{tab:linear_large}
		\begin{tabular}{@{}cccc@{}}
		\toprule
		\textbf{Method}                   & \textbf{$d$} & \textbf{SHD}    & \textbf{Runtime (hours)} \\ \midrule
		\multirow{5}{*}{\textbf{DAGMA}}   & 200           & 37.77$\pm$8.80   & 0.06$\pm$ 0.00       \\
		                                  & 300           & 86.65$\pm$19.63  & 0.13$\pm$ 0.00       \\
		                                  & 500           & 211.68$\pm$78.88  & 0.37$\pm$0.02       \\
		                                  & 800           & 285.90$\pm$56.89  & 1.15$\pm$0.06       \\
		                                  & 1000          & 473.70$\pm$101.63  & 2.09$\pm$0.08       \\\midrule
		\multirow{5}{*}{\textbf{GOLEM}}   & 200           & 215.68$\pm$66.91   & 0.32$\pm$0.00      \\
		                                  & 300           & 552.72$\pm$113.60   & 0.73$\pm$0.02      \\
		                                  & 500           & 1390.43$\pm$213.77  & 2.32$\pm$0.04      \\
		                                  & 800           & 2919.14$\pm$191.38 & 7.13$\pm$0.10      \\
		                                  & 1000          & 4083.03$\pm$157.58 & 12.50$\pm$0.22      \\\midrule
		\multirow{5}{*}{\textbf{NOTEARS}} & 200           & 105.58$\pm$24.19  & 1.28$\pm$0.14       \\
		                                  & 300           & 217.73$\pm$48.30  & 3.42$\pm$0.28     \\
		                                  & 500           & 441.72$\pm$93.01  & 10.84$\pm$0.62     \\
		                                  & 800           & 829.08$\pm$118.10 & 28.79$\pm$1.22     \\
		                                  & 1000          & $-$ & $> 36$   \\ \bottomrule
		\end{tabular}
		\end{table}
		
	\vspace{-0.2in}
	\subsubsection{DAGMA vs GOLEM in Sparser Graphs}
	\label{app:dagma_vs_golem}
		We note that in the work by \citet{Ng.2020}, the authors performed experiments on large number of nodes \textbf{only for ER2 graphs}, that is, sparser graphs.
		It was reported in \citet{Ng.2020} that their GOLEM method performed reasonably well	 for large number of nodes. However, as shown in Figure \ref{fig:linear_large}, for denser graphs such as ER4 and SF4, the performance of GOLEM degrades very fast as $d$ increases.
		In this section, we experiment with the same graph model as in \citep{Ng.2020}, i.e., ER2, for $d\in \{200,300,500,800,1000,2000\}$
		We note that even though GOLEM is competitive in this regime, DAGMA still obtains \emph{significant} improvements.
		
\begin{figure}[!t]
			\centering
			\includegraphics[width=.7\textwidth]{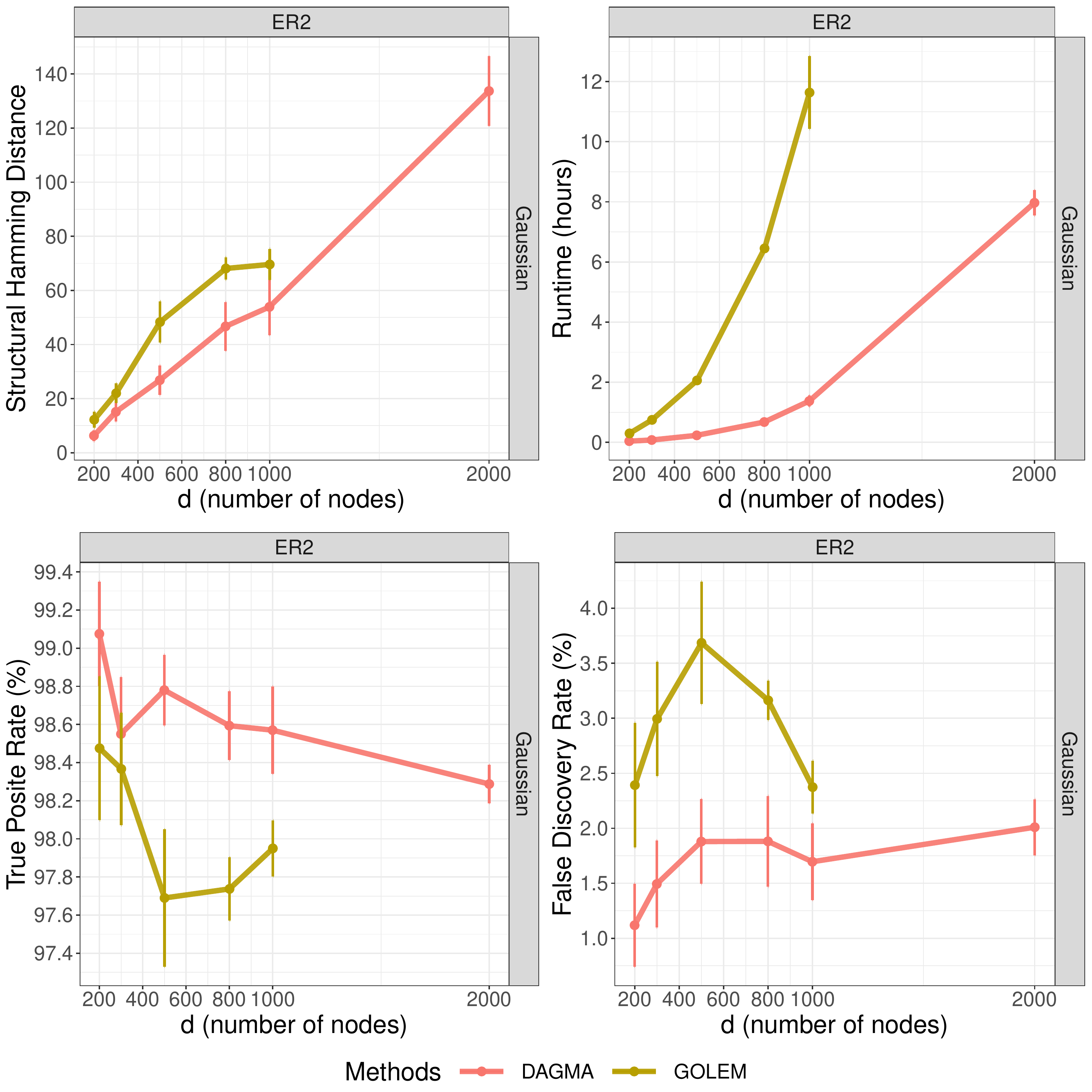}
			\caption{
				SHD, Runtime, TPR and FDR of DAGMA and GOLEM for a graph type ER2 with Gaussian noise.
				In all cases, lower is better \emph{except} for the TPR.
				Error bars represent standard errors over 10 simulations.
				More details are given in Section \ref{app:dagma_vs_golem}
			}
			\label{fig:linear_large_dagma_vs_golem}
		\end{figure}

		DAGMA was run under the same setting described in Section \ref{app:linear_small}. The results are reported in Figure \ref{fig:linear_large_dagma_vs_golem}.
		Here we note in particular that for $d=1000$, DAGMA obtains an \textbf{improvement} of $22.5\%$ in \textbf{SHD}, and runs $8.5$ times \textbf{faster} than GOLEM, even though the latter is customized for linear models.
		For $d=2000$, GOLEM took more than 36 hours per simulation, hence, we could not report its performance.
		
	\vspace{-0.15in}
	\subsubsection{DAGMA vs NOTEARS and GOLEM in Denser Graphs}
	\label{app:linear_denser}
\begin{figure}[!t]
		\centering
		\includegraphics[width=.65\textwidth]{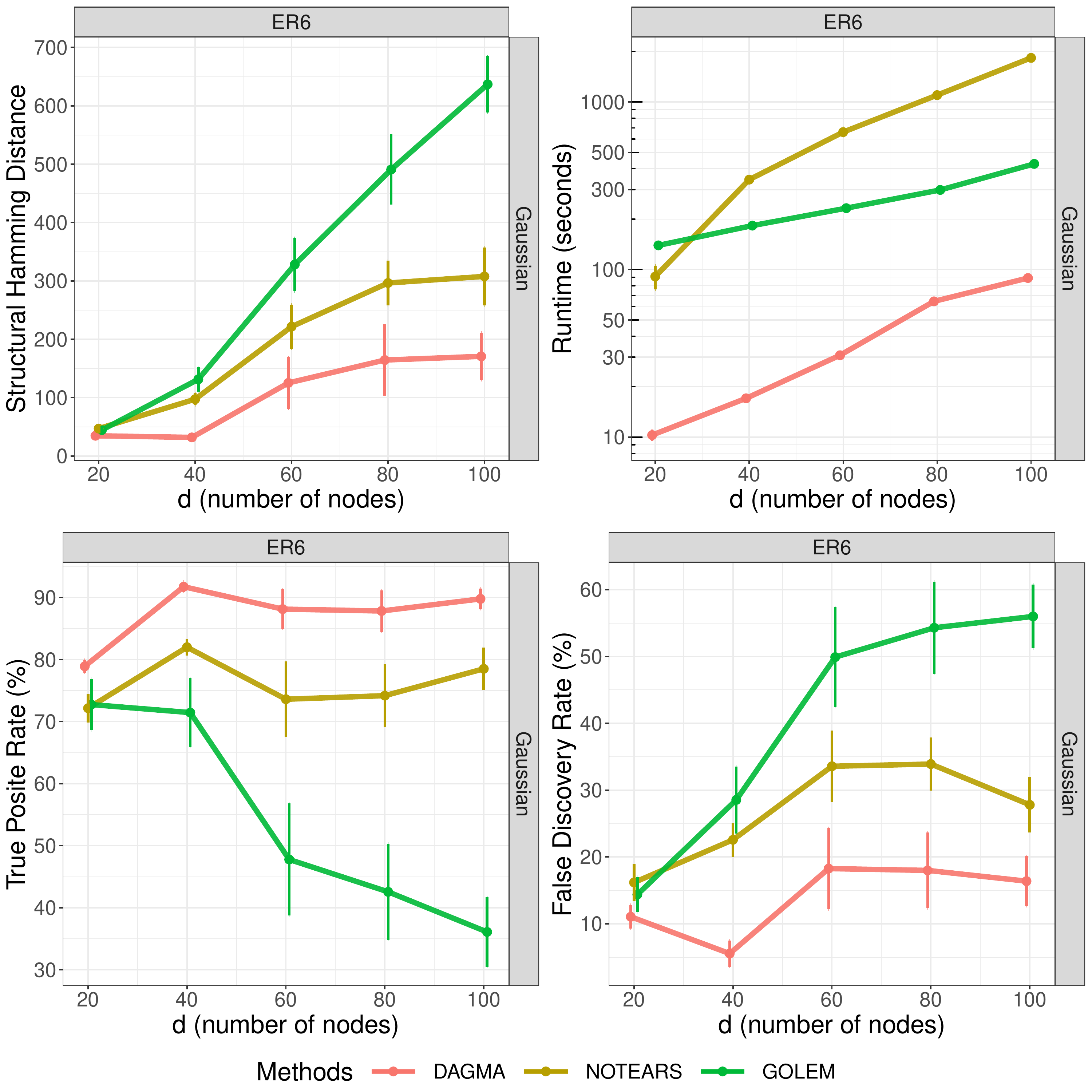}
		\caption{SHD, Runtime, TPR and FDR of DAGMA, GOLEM, and NOTEARS for a graph type ER6 with Gaussian noise.
				In all cases, lower is better \emph{except} for the TPR.
				Error bars represent standard errors over 10 simulations.
				More details are given in Section \ref{app:linear_denser}
			}
		\label{fig:linear_denser}
	\end{figure}

		\vspace{-0.05in}
		For completeness, we run experiments on a denser graph type such as ER6 with Gaussian noise, for $d \in \{20, 40, 60, 80, 100\}$.
		DAGMA was run under the same setting described in Section \ref{app:linear_small}. The results are reported in Figure \ref{fig:linear_denser}.
		Here we note in particular that for $d=100$, DAGMA obtains an \textbf{improvement} of $73.1\%$ and $44.5\%$ in \textbf{SHD} against GOLEM and NOTEARS, respectively; also, DAGMA runs $4.8$ and $20.6$ times \textbf{faster} than GOLEM and NOTEARS, respectively.
		Finally, we note that even in the regime of small number of variables, GOLEM's performance degrades very fast for denser graphs, while DAGMA's performance remains the best among the three methods.
	
\subsection{SEM: Non-Linear Setting}
\label{app:nonlinear_models}

	\subsubsection{Logistic Model}
	\label{app:nonlinear_logistic}
		Given the ground-truth DAG $B \in \{0,1\}^{d\times d}$ from an ER4 graph, we assigned edge weights independently from $\mathrm{Unif}\left([-2,-0.5]\ \union\ [0.5,2]\right)$ to obtain a weight matrix $W \in \sR^{d\times d}$. 
		Given $W$, we sampled $X_j = \text{Bernoulli}(\operatorname{exp} (w_j^\top X)/(1+\operatorname{exp} (w_j^\top X))), \forall j \in [d]$.
		Based on this model, we generated random datasets $\mX \in \sR^{n\times d}$ by generating the \iid rows.
		For each simulation, we generated $n = 5000$ samples for graphs with $d \in\{ 20,40,60,80,100,200,400,600,800,1000\}$ nodes. 
		
		To measure the quality of a model, we use the log-likelihood loss 
		\begin{align}
		\label{eq:logloss}
			Q(f,\rmX)  = \frac{1}{n}\sum_{i=1}^d\mathbf{1}_n^\top\left(\log (\mathbf{1}_n+\text{exp}(f_i(\rmX)))-\mathbf{x}_i\circ f_i(\rmX)\right).
		\end{align}
		
		The implementation details of the baselines are listed below: 
		\begin{itemize}
		    \item GES (specifically, the FGES algorithm in \citep{ramsey2017}) and PC \citep{Spirtes.2000} are standard baselines for structure learning. Their implementation is based on the \texttt{py-causal} package, available at \url{https://github.com/bd2kccd/py-causal}. The exact set of hyperparameters used are:
		    \begin{itemize}
		    	\item For PC: \texttt{testId = `disc-bic-test', depth = 4, fasRule = 2, dataType = `discrete',  conflictRule = 1, 
	        		concurrentFAS = True,
	 		       useMaxPOrientationHeuristic = True}.
		    	\item For GES: \texttt{scoreId = `bdeu-score', maxDegree = 5, dataType = `discrete',\\ faithfulnessAssumed = False}.
		    \end{itemize}
		    \item The NOTEARS method in \citet{Zheng.2018jsc} was implemented using the author's Python code available at: \url{https://github.com/xunzheng/notears}. Its score function is also the log-likelihood loss as defined in eq.\eqref{eq:logloss}. For the $\ell_1$ coefficient, for a fair comparison, we use the same value used for DAGMA. For the rest of hyperparameters, we use their default values.
		\end{itemize}
		
		We use the following setting for DAGMA (Algorithm \ref{algo:path_following}): Number of iterations $T=4$, initial central path coefficient $\mu^{(0)} = 10$, decay factor $\alpha = 0.1$, $\ell_1$ coefficient $\beta_1 = 0.01$, log-det parameter $s=\{1, .9, .8, .7\}$.
		For each problem in line 3 of Algorithm \ref{algo:path_following}, we implement an adaptive gradient method using the ADAM optimizer \citep{kingma2014adam}.
		The hyperparameters for ADAM are: Learning rate of $3\times 10^{-4}$, and $(\beta_1,\beta_2) = (0.99,0.999)$.
		For $t = \{0,1,2\}$, we run ADAM for $10^4$ iterations or until the loss converges, whichever comes first.
		For $t = 3$, we run ADAM for $5\times 10^4$ iterations or until the loss converges, whichever comes first.
		We consider that the loss converges if the relative error between subsequent iterations is less than $10^{-6}.$
		Finally, as in \citep{Zheng.2018jsc,Zheng.2020,Ng.2020}, a final thresholding step is performed as it was shown to help reduce the number of false discoveries. For all cases, we use a threshold of $0.3$.
		
\begin{figure}[!t]
			\centering
			\includegraphics[width=.65\textwidth]{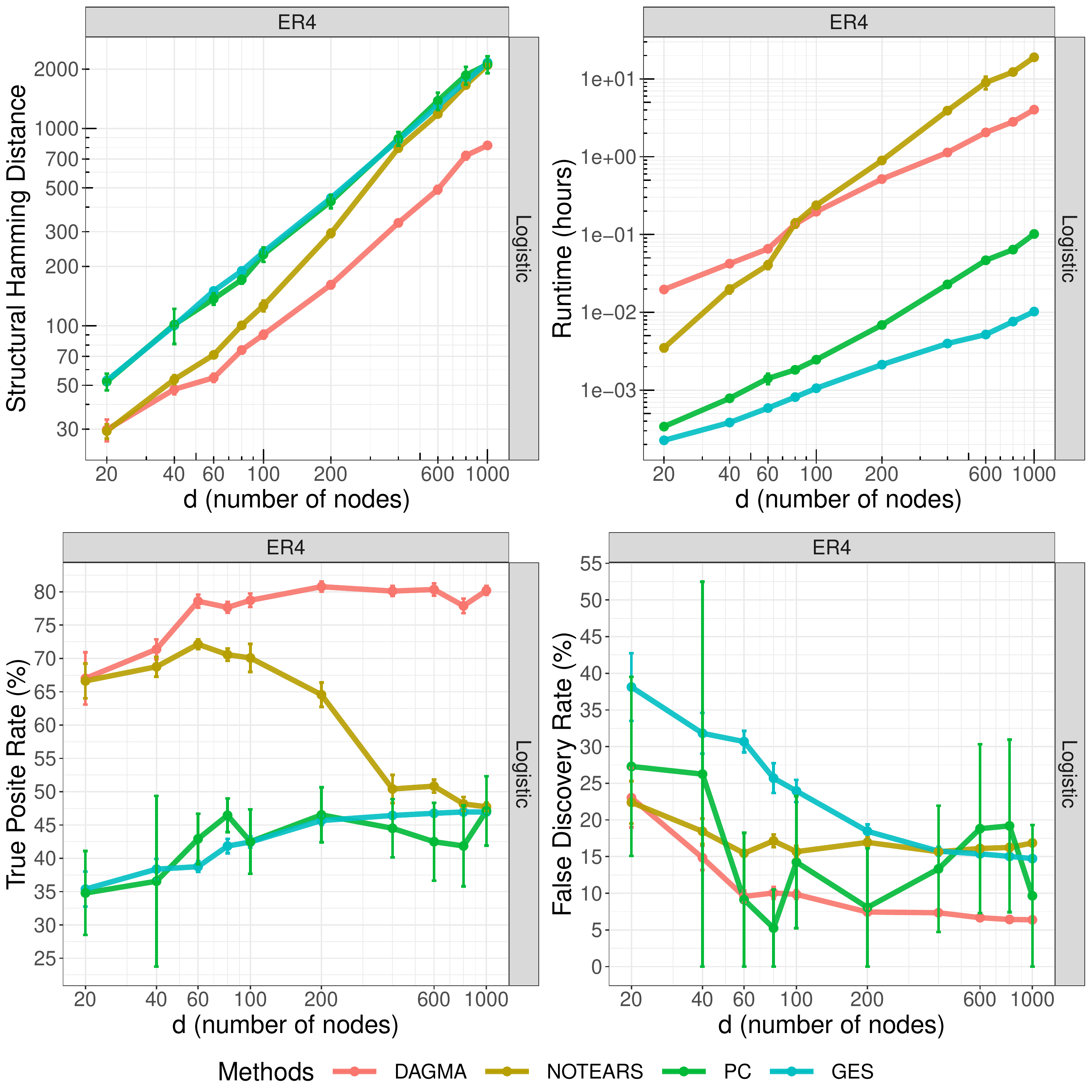}
			\caption{SHD, Runtime, TPR and FDR of all methods for a graph type ER4 and logistic model.
				In all cases, lower is better \emph{except} for the TPR.
				Error bars represent standard errors over 10 simulations.
				More details are given in Section \ref{app:nonlinear_logistic}
				}
			\label{fig:nonlinear_logistic}
		\end{figure}

		The results are shown in Figure \ref{fig:nonlinear_logistic}.
		We note that for $d=1000$, DAGMA obtains an \textbf{improvement} of $60\%$ in \textbf{SHD} and runs $4.8$ times \textbf{faster} than NOTEARS.
		Finally, we note that GOLEM is not considered for the nonlinear models as it only works for linear ones.
		
	\subsubsection{Neural Network Model}
	\label{app:nonlinear_mlp}
		We mainly follow the nonlinear setting of \citet{Zheng.2020}. 
		That is, given a ground-truth graph $G$, we simulate the SEM:
		\[
			X_j = f_j(X_{\text{pa}(j)}) + Z_j, \forall j\in [d],
		\]
		where $Z_j\sim \mathcal{N}(0,1)$ is a standard Gaussian noise. 
		Here $f_j$ is a randomly initialized multilayer perceptron (MLP) with one hidden layer of size 100 and sigmoid activation.
		Similar to previous experiments, we generate a dataset $\mX \in \sR^{n\times d}$, with $n=1000$ \iid samples.
		
		For this setting, we only compare to NONLINEAR NOTEARS \cite{Zheng.2020}.
		We refer the reader to \citep{Zheng.2020} for a comprehensive comparison with other baselines.
		For NONLINEAR NOTEARS and DAGMA, each $f_\theta$ is modeled by a MLP with one hidden layer of size $10$ and sigmoid activation. 
		In contrast to the original implementation of NONLINEAR NOTEARS \citep{Zheng.2020} which uses the square loss, we use the log-likelihood as in \cite{buhlmann2014cam} as we observe better performances for both methods.
		
		We use the following setting for DAGMA (Algorithm \ref{algo:path_following}): Number of iterations $T=4$, initial central path coefficient $\mu^{(0)} = 0.1$, decay factor $\alpha = 0.1$, $\ell_1$ coefficient $\beta_1 = 0.02$, log-det parameter $s=1$.
		For each problem in line 3 of Algorithm \ref{algo:path_following}, we implement an adaptive gradient method using the ADAM optimizer \citep{kingma2014adam}.
		The hyperparameters for ADAM are: Learning rate of $2\times 10^{-4}$, and $(\beta_1,\beta_2) = (0.99,0.999)$.
		For $t = \{0,1,2\}$, we run ADAM for $7\times 10^4$ iterations or until the loss converges, whichever comes first.
		For $t = 3$, we run ADAM for $8\times 10^4$ iterations or until the loss converges, whichever comes first.
		We consider that the loss converges if the relative error between subsequent iterations is less than $10^{-6}.$
		Finally, as in \citep{Zheng.2018jsc,Zheng.2020,Ng.2020}, a final thresholding step is performed as it was shown to help reduce the number of false discoveries. For all cases, we use a threshold of $0.3$.
		
\begin{figure}[!t]
			\centering
			\includegraphics[width=.7\textwidth]{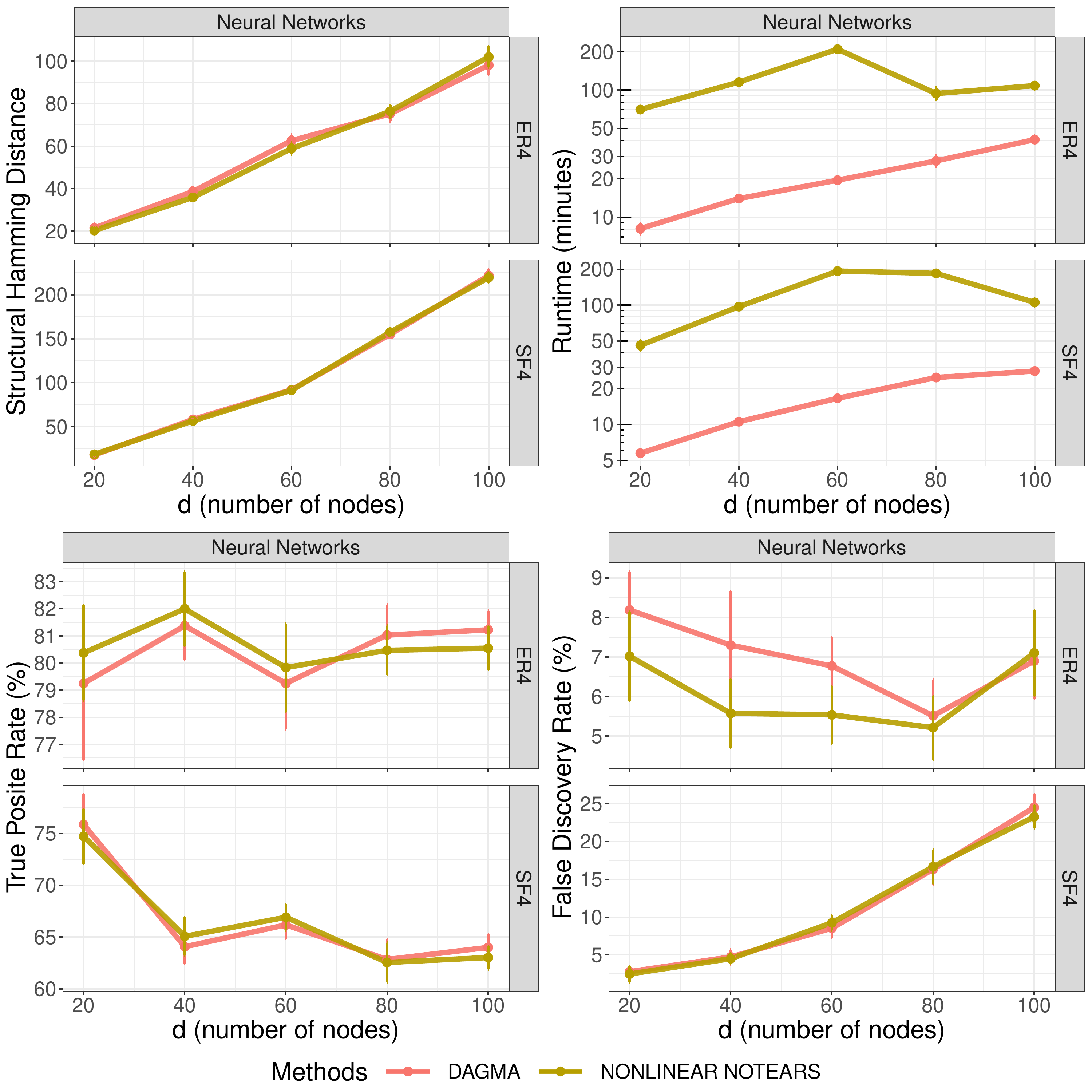}
			\caption{SHD, Runtime, TPR and FDR of all methods for a graph type ER4 and logistic model.
				In all cases, lower is better \emph{except} for the TPR.
				Error bars represent standard errors over 10 simulations.
				More details are given in Section \ref{app:nonlinear_mlp}
				}
			\label{fig:nonlinear_mlp}
		\end{figure}

		The results are shown in Figure \ref{fig:nonlinear_mlp}.
		We note that DAGMA and NOTEARS obtain similar performances in SHD; however, DAGMA can obtain 3x to 10x speedups over NONLINEAR NOTEARS.
		Finally, we note that GOLEM is not considered for the nonlinear models as it only works for linear ones.
		
\section{Broader Impacts} 
\label{sec:broader_impacts}   
	A potential misuse of this type of work would be to purposely (or not) run the method proposed on a dataset that is biased. 
	Since we do not formally deal with inherent biases in the dataset (e.g., unfairness due to selection bias), it is possible to learn relationships that are not present in reality.
	A user can then (un)intentionally report a result incorrectly claiming to have found the cause of a certain variable, thus, creating misinformation.

\end{document}